\documentclass[journal]{IEEEtai}

\usepackage[colorlinks,urlcolor=blue,linkcolor=blue,citecolor=blue]{hyperref}

\usepackage{color,array}
\usepackage{amsmath}
\usepackage{algorithm}
\usepackage{algpseudocode}

\usepackage{graphicx}
\usepackage[justification=centering]{caption}

\begin{document}

\title{Efficient Model Adaptation for Continual Learning at the Edge} 

\author{Zachary A. Daniels, Jun Hu, Michael Lomnitz, Phil Miller, Aswin Raghavan, Joe Zhang, Michael Piacentino, and David Zhang 
\thanks{Submitted: May 31, 2023}
\thanks{This research is based upon work supported in part by the Office of the Director of National Intelligence (ODNI), Intelligence Advanced Research Projects Activity (IARPA), via Contract No: 2022-21100600001. The views and conclusions contained herein are those of the authors and should not be interpreted as necessarily representing the official policies, either expressed or implied, of ODNI, IARPA, or the U.S. Government. The U.S. Government is authorized to reproduce and distribute reprints for governmental purposes notwithstanding any copyright annotation therein.}
\thanks{All authors are researchers with the Center for Vision Technologies, SRI International, Princeton, New Jersey 08540 USA (corresponding author email: zachary.daniels@sri.com)}
}


\maketitle

\begin{abstract}
Most machine learning (ML) systems assume stationary and matching data distributions during training and deployment. This is often a false assumption. When ML models are deployed on real devices, data distributions often shift over time due to changes in environmental factors, sensor characteristics, and task-of-interest. While it is possible to have a human-in-the-loop to monitor for distribution shifts and engineer new architectures in response to these shifts, such a setup is not cost-effective. Instead, non-stationary automated ML (AutoML) models are needed. This paper presents the Encoder-Adaptor-Reconfigurator (EAR) framework for efficient continual learning under domain shifts. The EAR framework uses a fixed deep neural network (DNN) feature encoder and trains shallow networks on top of the encoder to handle novel data. The EAR framework is capable of 1) detecting when new data is out-of-distribution (OOD) by combining DNNs with hyperdimensional computing (HDC), 2) identifying low-parameter neural adaptors to adapt the model to the OOD data using zero-shot neural architecture search (ZS-NAS), and 3) minimizing catastrophic forgetting on previous tasks by progressively growing the neural architecture as needed and dynamically routing data through the appropriate adaptors and reconfigurators for handling domain-incremental and class-incremental continual learning. We systematically evaluate our approach on several benchmark datasets for domain adaptation and demonstrate strong performance compared to state-of-the-art algorithms for OOD detection and few-/zero-shot NAS.
\end{abstract}

\begin{IEEEImpStatement}
Generally, ML systems assume stationary and matching data distributions during training and deployment. In practice, deployed ML systems encounter shifts in the input distribution over time, e.g., due to environmental factors. Human-in-the-loop monitoring for distribution shifts and hand-engineering new ML architectures is cost-prohibitive. Our framework automatically identifies when domain shifts occur and adapts the neural architecture to account for these shifts. In edge computing and Internet-of-Things (IoT) applications, the deployed hardware has limited compute resources. Compared to many existing approaches for AutoML, we focus on fast, computationally-inexpensive methods for OOD detection and NAS to learn low-parameter adaptor models.
\end{IEEEImpStatement}
\begin{IEEEkeywords}
AutoML, Continual Learning, Edge Computing, Out-of-Distribution Sample Detection, Progressive Neural Networks, Zero-Shot Neural Architecture Search
\end{IEEEkeywords}

\section{Introduction}


\IEEEPARstart{I}{n} traditional ML, it is assumed that the distribution of input features and output labels do not change once the model is trained; i.e., the model is fit to a stationary distribution of features on a specific task, and during inference/deployment, the model is applied to the same task with a matching distribution of input features. In contrast, ML models deployed on real-world devices for applications involving IoT, edge computing, and sensor network analysis often contend with distribution shift over time due to changes in 1) the sensor (e.g., a model trained on high-resolution imagery is applied to low-resolution imagery), 2) the task space (e.g., a model trained to detect one set of vehicles is reused for detecting a different set of vehicles), or 3) the environment (e.g., a model trained on sunny days is run during a storm).

While it is possible for a human-in-the-loop to monitor for distribution shifts and engineer new architectures to account for these shifts, such a setup is generally not cost-effective. A more practical alternative is to develop a ML system that can automatically determine 1) when the distribution of inputs and outputs has changed, and 2) determine how to adapt its architecture to handle the new data distribution while maintaining performance on previous distributions. Furthermore, for many of the applications involving ML under distribution shifts, adaptation must be done quickly on resource-constrained hardware. This work proposes an approach to efficient non-stationary AutoML.

Specifically, we focus on the problem of domain- and class-incremental continual learning \cite{de2021continual, van2022three} where distribution shifts in the input and output spaces occur over time. Domain-incremental continual learning involves learning the same kind of problem under different contexts (e.g., modalities). Class-incremental continual learning involves incrementally learning to assign labels to a growing set of classes, potentially under different contexts. 

In our experiments, we consider four cases of domain shift: 1) in the most extreme case, both the input modality and output label sets change, 2) the input modality changes, but the class labels remain the same, 3) the input modality remains the same, but the label sets change, and 4) in the most subtle case, the input modality and class labels remain the same, but environmental factors change.

The goal of this work is to learn to identify when the data distribution has changed and rapidly adapt ML models to new domains (non-stationary AutoML) with limited compute resources. We introduce the Encoder-Adaptor-Reconfigurator (EAR) framework, which consists of three components:
\begin{itemize}
    \item \textbf{Encoder: } Fixed pre-trained feature extraction backbone
    \item \textbf{Adaptors: } Shallow NNs that facilitate feature transfer to new data distributions
    \item \textbf{Reconfigurator: } Light-weight model that enables rapid adaptation to new task spaces with little re-training
\end{itemize}
We focus on the research problems of 1) how to identify when data distributions shift, 2) how to grow the ML model as needed, balancing performance on the new domain and model efficiency, and 3) how to perform continual learning via intelligent dynamic data routing through adaptors/reconfigurators to minimize catastrophic forgetting of previous domains. This paper presents the following contributions:
\begin{itemize}
    \item Introduction of the Encoder-Adaptor-Reconfigurator framework for efficient model adaptation to distribution shifts on resource-constrained hardware
    \item Formulation of and training procedure for learning deep hyperdimensional (HD) \cite{kanerva2009hyperdimensional} adaptor-reconfigurators for joint OOD detection and robust classification
    \item Formulation of a spectral analysis-based approach to zero-shot neural architecture search \cite{mellor2021neural}
    \item Demonstration of the EAR framework in a continual learning setting via progressive neural networks \cite{rusu2016progressive,fayek2020progressive} with dynamic data routing
    \item High performance compared to state-of-the-art algorithms for OOD detection and few/zero-shot NAS on benchmark domain adaptation datasets
\end{itemize}

\section{Background}
\subsection{Progressive Neural Networks for Continual Learning}
We explore the problem of continual learning \cite{de2021continual,van2022three} where an agent is trained on a sequence of tasks, and the agent must balance plasticity vs stability: it must learn to solve the new task while minimizing (catastrophic) forgetting of previous tasks. Van de Ven et al. \cite{van2022three} categorized continual learning into three types: task-, domain-, and class-incremental. Task-incremental learning involves solving sequences of tasks where the agent is explicitly told the current task-of-interest. In the domain-incremental setting, the agent does not know what the the current task-of-interest is, but the structure of the problem does not change between tasks (only the input distribution exhibits shifts), and the agent can solve the current task without explicitly identifying the task. In the class-incremental setting, the structure of the problem changes over time (i.e., new class sets are added). In this case, the agent must both identify and subsequently solve the current task-of-interest. The EAR framework is designed for the most-challenging class-incremental setting, but can be trivially applied to the task-incremental and domain-incremental settings.

There are three broad approaches to training continual learning agents \cite{chen2018lifelong}. Replay-based methods save examples of previously encountered tasks (explicitly, via exemplars/prototypes, via generative memory, or in compressed representations), and periodically replay these examples to minimize catastrophic forgetting. Regularization-based methods impose constraints on the learning behavior of the network (e.g., through modification of the loss function) to prevent the model from overfitting to the new task and overwriting knowledge about the old tasks. Architecture-based methods intelligently grow and prune the model architecture. Our EAR framework is an architecture-based approach to continual learning. Other architecture-based approaches include Progressive Neural Networks \cite{rusu2016progressive,fayek2020progressive}, Dynamically Expanding Networks \cite{yoon2017lifelong}, ``Compacting, Picking, and Growing'' \cite{hung2019compacting}, ``Learn to Grow'' \cite{li2019learn}, and Comprehensively Progressive Bayesian NNs \cite{yang2022robust}.

The EAR architecture can be thought of as a special case of a progressive NN. Progressive NNs grow lateral connections, thus, avoiding forgetting at the cost of increased resource use. The EAR framework progressively grows adaptors and reconfigurators off of a frozen feature encoder backbone. To extend progressive NNs, we propose a novel method for dynamically routing data through the appropriate adaptors/reconfigurators, and we utilize ZS-NAS to identify where the adaptors should be added and the structure of the adaptors.

\subsection{Out-of-Distribution Detection with Deep Neural Networks}
Our approach aims to automatically identify when the input distribution has changed (OOD detection) \cite{yang2021generalized}. In particular, we focus on the setting of novelty detection \cite{markou2003novelty} where the OOD detector only sees in-distribution (ID) samples during training. Our model learns adaptors that project data into learned representations that can be used for joint OOD detection and classification. There are multiple approaches to OOD detection with deep networks. In the most basic approach, a deep network is trained with standard cross-entropy and feature vector immediately preceeding the classifier layer is used to train an OOD detection model (e.g., using off-the-shelf classic OOD and novelty detection algorithms such as the one-class SVM \cite{manevitz2001one}, empirical and robust covariance estimates \cite{pena2001multivariate}, local outlier factors \cite{breunig2000lof}, isolation forests \cite{liu2008isolation,liu2012isolation}). Alternatively, statistics about the softmax probability distribution output by a pre-trained neural network model can be analyzed to separate OOD from ID samples (e.g., \cite{hendrycks2016baseline}). Extending beyond single-layer approaches, state-of-the-art methods fuse information from multiple layers of pre-trained networks to detect OOD samples (e.g., gram matrices \cite{sastry2020detecting} and hyperdimensional feature fusion \cite{wilson2023hyperdimensional}). Finally, the structure and training of the NN can be modified to learn representations designed for joint OOD detection and classification, e.g., using supervised contrastive learning in combination with a deep nearest neighbor classifier \cite{sun2022out}. Our approach combines multi-layer feature analysis with learned representations, extending the HD feature fusion method by using adaptors to automatically learn projections from features to HD vectors instead of using random fixed projection matrices.

\subsection{Hyperdimensional Computing}
Our approach to OOD detection and classification is built around hyperdimensional computing (HDC) \cite{kanerva2009hyperdimensional, thomas2021theoretical, kleyko2023survey}. HDC is a neuro-inspired neurosymbolic compute paradigm that represents discrete pieces of information as high-dimensional, low-precision, distributed vectors. In contrast to DNNs, HDC uses low-power, requires low-precision, and has been shown to be robust to corruptions in the input data. HDC is built off of the mathematics of manipulating random pseudo-orthogonal vectors in high-dimensional spaces. HDC is built around two key operations: 1) \textit{binding} which takes two input vectors and generates a new vector that is dissimilar to each of the inputs and 2) \textit{bundling} (a.k.a. superposition) which takes two or more input vectors and generates a new vector that is similar to the inputs. We learn adaptors that represent layer-wise features as pseduo-orthogonal binary vectors, and then, reconfigurators use majority voting to bundle all of the samples from a class into a binary class prototype vector. New samples can be classified into existing classes if its HD vector is close to one of the class prototype vectors based on hamming distance, or it can be classified as OOD if it is far from all of the class prototypes.

\subsection{Zero-Shot Neural Architecture Search}
DNNs have been shown to naturally facilitate transfer learning \cite{zhuang2020comprehensive}, even by simply finetuning the final layer(s) of the model. However, when there are dramatic shifts in the input distribution or task-space, finetuning just the final layers may be insufficient, and finetuning the entire network could be cost-prohibitive and lead to catastrophic forgetting of previous domains. Furthermore, a network architecture which is optimized to one domain may not perform optimally on another. The EAR framework freezes a feature encoder network trained on one domain and learns sets of adaptors/reconfigurators that tap the encoder at different locations. To optimize performance on the new task while constraining the growth of the model, it is necessary to carefully design the adaptor layers and identify which tap points will best facilitate transfer learning to the new domain. To identify the location and structure of the adaptors and reconfigurators, we employ NAS \cite{elsken2019neural}.

NAS consists of three components:
\begin{itemize}
\item The \textbf{search space} is the space of what architectures can be represented in the search process
\item The \textbf{search strategy} is the algorithm used to find and test architectures, balancing exploration and exploitation. Examples include reinforcement learning, Bayesian optimization, and evolutionary algorithms.
\item The \textbf{performance estimation strategy:} quantifies how good an architecture is, what the performance measure is, and what the constraints are.
\end{itemize}
In this work, the search space defines what the adaptors look like and where to place them, the search strategy compatible with our EAR framework is any global optimization strategy (we use Bayesian optimization), and our main novel contribution is defining the performance estimation strategy. 

The EAR framework is intended to be deployed on low-resource hardware and ideally adapts in a quick and efficient manner. For these reasons, we focus on zero-shot neural architecture search \cite{mellor2021neural,abdelfattah2021zero,chen2021bench}. ZS-NAS evaluates candidate architectures without training the architecture via proxy heuristics, which predict properties correlated with how well a candidate model is expected to perform after it is trained. We propose an approach based on spectral analysis of the feature spaces of the adaptors, but a number of other ZS-NAS approaches exist, including snip \cite{lee2018snip}, grasp \cite{wang2020picking}, fisher score \cite{turner2019blockswap}, Jacobian-covariance score \cite{mellor2021neural}, Synflow \cite{tanaka2020pruning}, grad-norm, $\Phi$-Score \cite{lin2021zen}, and Zen-NAS \cite{lin2021zen}. Most work formulates the proxy heuristic in terms of the gradients of the candidate architecture over a random batch of data. In contrast, our approach uses a gradient-free proxy, making it more computationally-efficient to compute.

\section{Methodology}
\subsection{The Encoder-Adaptor-Reconfigurator Framework for Continual Learning}
\begin{figure}[t]
{\centering
\includegraphics[width=0.52\linewidth]{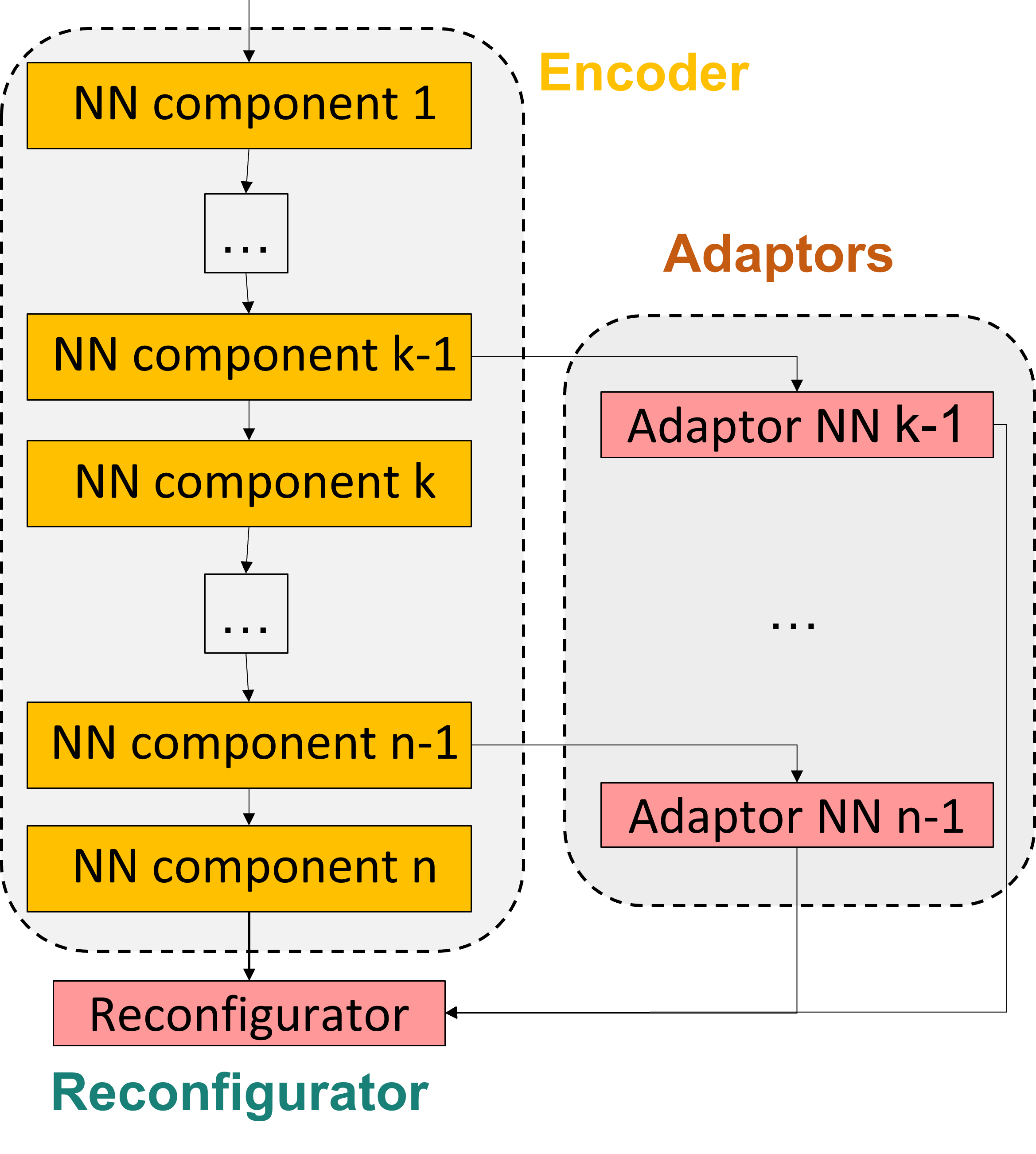}
\caption{High-level diagram of the EAR architecture}
\label{fig:ear_high_level}
}
\end{figure}
\begin{figure*}[t]
{\centering
\includegraphics[width=0.85\linewidth]{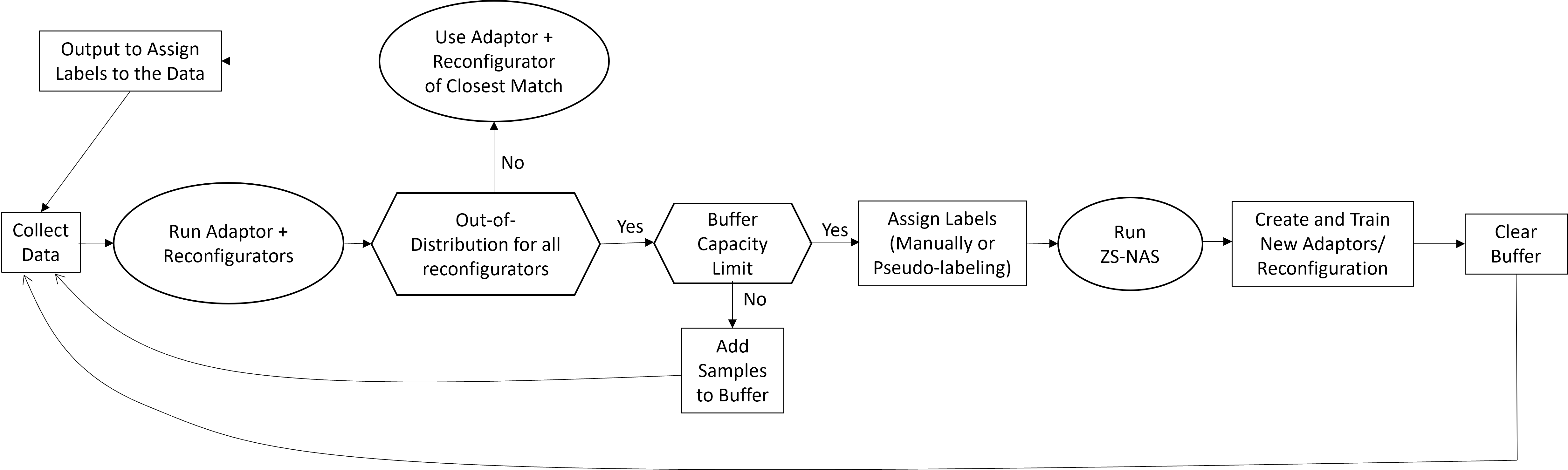}
\caption{Control flow for continual learning using the EAR framework}
\label{fig:ear_continual}
}
\end{figure*}
We propose the EAR framework for handling domain shift over time in compute-constrained settings (Fig. \ref{fig:ear_high_level}). The framework is composed of three components: a fixed pre-trained feature extraction backbone \textit{encoder}, a set of shallow NN \textit{adaptors} that facilitate feature transfer to new data distributions, and a light-weight \textit{reconfigurator} that enables rapid adaptation to new task spaces with little re-training.

In our experiments, the encoder is a DNN pre-trained on a large-scale dataset and finetuned on the first domain encountered; after which, it is frozen. As the model encounters new domains, shallow adaptors, which are laterally connected to tap points of the encoder, are learned. These adaptors efficiently transform the features tuned for the first domain to be useful for subsequent domains. Our adaptors consist of a few convolutional and dense layers that transform unconstrained feature vectors into binary HD feature vectors (Section \ref{sec:hd_ood}). The output per-adaptor HD vectors feed into a reconfigurator that is the predictive model for the new domain (a joint OOD detector and classifier). The reconfigurator bundles all of per-adaptor HD vectors into a single aggregated HD vector per input instance. During training, the aggregated HD vectors of all data in the training set from a single class are bundled into a single prototype per class. During inference, classification is performed by comparing the aggregated HD vector of an instance with all of the class prototype vectors stored by the reconfigurator. The instance is assigned the class of the nearest prototype, or if it isn't close to any prototype, it is assigned to be OOD for the domain associated with the reconfigurator (details in Section \ref{sec:hd_ood}).

In Fig. \ref{fig:ear_continual}, we show the high-level control flow of the EAR architecture in a continual learning scenario. We assume that the model only sees data streaming from a single domain/task at a given time, but the domain/task can shift at any time. Thus, the model needs to operate over sequences of tasks where task boundaries are not known. As new data arrives, it is passed through the encoder once and then passed through each set of adaptors/reconfigurators. If the data is ID according to any reconfigurator, then it is classified according to the reconfigurator of closest match (smallest OOD score). When a new task is encountered, new samples will appear to be OOD to all reconfigurators. Once the model is sufficiently confident that there has been a domain shift, it will verify with an oracle (e.g., human-in-the-loop) that a shift has occurred, and it will start collecting data for the new domain. Note that we do not discuss how to automatically annotate novel samples in this paper as it is not our focus, and thus assume an oracle exists for this purpose. Once the buffer hits its capacity limit, the oracle labels the data, and the system uses ZS-NAS (Section \ref{sec:zsnas}) to identify the structure and placements of a new set of adaptors and reconfigurator, trained on the collected data. To determine when a new domain/task appears, the model monitors whether the proportion of the last $N$ data samples assigned as OOD is greater than a specified threshold; after which, the update process is triggered.  

\subsection{Deep Hyperdimensional Computing for Joint Out-of-Distribution Sample Detection and Classification}
\label{sec:hd_ood}
Wilson et al. showed that HDC \cite{wilson2023hyperdimensional} can be combined with pre-trained DNNs for novelty detection (i.e., when the OOD detector only sees ID data during training). Their approach projects the outputs of every layer of a pre-trained NN to a high-dimensional random vector via a binding operation. The layer-wise HD vectors are aggregated into a single vector, and HD prototype vectors are learned for each class. If the HD vector of a new instance has a distance larger than a fixed threshold to all of the class prototypes, it is flagged as OOD.

This work extends the method of Wilson et al. The novelty of our approach is: 1) instead of using random fixed linear projections to map layer-wise outputs to HD vectors, we use adaptors to learn non-linear projections to a pre-determined fixed set of HD vectors; 2) we perform fusion from a small subset of layers determined via ZS-NAS instead of every layer; 3) we project to binary HD vectors instead of real-valued HD vectors, saving memory and enabling lighter computation; and 4) our method enables the model to learn to perform joint OOD detection-classification in an end-to-end manner, resulting in more discriminative representation learning.


\subsubsection{Basic Model for Classification}
To begin, we explain how the adaptors can learn a classification model by mapping input samples to HD vectors using a combination of NNs and HDC operations. For a specific domain, given an input sample $x$, an adaptor $f_{ada}^{tap}(.)$ map features from a tap point $tap$ of the encoder model $f_{enc}(.)$ to a binary HD vector $h_{ada}^{tap}(.)$:
\begin{equation}
\begin{split}
    \pi_{ada}^{tap}(x) = f_{ada}^{tap}(f_{enc}(x)), \\
    h_{ada}^{tap}(x) = \text{sample element } i \\
    \text{ as 1 with probability } \pi_{ada}^{tap, \; (i)}(x), \\
    \text{as 0 with probability } 1-\pi_{ada}^{tap, \; (i)}(x)\\
    \forall i \in [0, len(h_{ada}^{tap}(x))]
\end{split}
\label{eq:adaptors}
\end{equation}
The adaptors predict a score between 0 and 1 for each element of the HD vector (e.g., via a sigmoid activation). To cast this pseudo-binary output vector to a binary HD vector, we sample each element according to the element-wise score.

The reconfigurator serves two purposes:
\begin{enumerate}
    \item It aggregates the HD vectors over the set of all adaptors $\mathcal{A}$ for the domain into an HD vector $h_{agg}(.)$ by bundling:
    \begin{equation}
    h_{agg}(x) = round\left(\frac{1}{|\mathcal{A}|}\sum_{(tap,ada) \in \mathcal{A}} h_{ada}^{tap}(x)\right)
    \label{eq:reconfigurator}
    \end{equation}
    \item It learns prototypes for each class $h_{proto}^{class}$ for the domain by bundling all training instances from a class $X_{C}$:
    \begin{equation}
    h_{proto}^{class} = round\left(\frac{1}{|X_{C}|}\sum_{x_{c} \in X_{C}} h_{agg}(x_{c})\right)
    \label{eq:prototypes}
    \end{equation}
\end{enumerate}
To assign a class label to an input instance $x$, we compute $h_{agg}(x)$ and select the class with smallest hamming distance:
\begin{equation}
\hat{y}(x) = argmin_{class} \; d_{hamming}(h_{agg}(x), h_{proto}^{class})
\label{eq:classification}
\end{equation}

\subsubsection{Training the Adaptors}
To train the adaptors, we first generate a unique HD vector per class per adaptor (i.e., for ten classes and five adaptors, fifty pseudo-orthogonal vectors are generated) via Algorithm \ref{alg:binary_HD} ($\otimes$ is the Kronecker product).

\begin{algorithm}
\caption{Algorithm for generating binary HD vectors}\label{alg:binary_HD}
\begin{algorithmic}
\Require $n > k$ 
\Require $n$ is power-of-two
\State $k \gets $ number of classes $\times$ number of adaptors
\State $n \gets $ dimensionality of HD vector + 1 
\State $C \gets \left[\begin{matrix} 1 & 1 \\ 1 & -1 \end{matrix}\right]$
\State $C_{0} \gets clone(C)$
\State $i = 0$
\While{$i < \log_{2}(n / 2)$}
\State $C \gets C_{0} \otimes C$
\State $i \gets i + 1$
\EndWhile
\State $C \gets C[1:n, 1:n]$
\State Shuffle the rows of $C$
\State $C \gets C[0:k, 0:n-1]$
\State $C[C=-1] \gets 0$
\end{algorithmic}
\end{algorithm}

Algorithm \ref{alg:binary_HD} first generates a symmetric orthogonal matrix satisfying $\{-1, 1\}^{n \times n}$. The first row and column of the matrix are removed to improve stability of training the adaptors. While this breaks orthogonality, the matrix remains pseudo-orthogonal. The rows of the matrix are shuffled (helps stabilize the training of the adaptors), and HD vectors for each combination of adaptor and class label are selected. Finally, all elements of the HD vectors that are $-1$ are set to $0$. 

We select the dimensionality of the HD vectors to be $2^{\lceil\log_{2}(\#adaptors * \#classes + 1)\rceil} - 1$. This ensures that for each adaptor-class pair, there will be a unique mutually (pseudo-)orthogonal HD vector. This vector serves as the target output for any data sample of the specific class that passes through the corresponding adaptor. By forcing orthogonality between the HD vectors across adaptors by construction of the target HD vectors, binding operations are not needed during aggregation.

To map from inputs $x$ to the target HD vectors for an adaptor, we treat the mapping as a high-dimensional binary multi-label classification problem. We use the weighted binary focal cross-entropy loss \cite{lin2017focal} averaged over the every element $i$ of the predicted HD vector, which forces the adaptors to focus on harder-to-classify samples during training:
\begin{equation}
    \ell_{foc} = \frac{1}{\#dims}\sum_{i=0}^{\#dims} -\alpha_i * (1-p^{(i)}_{target})^{\gamma} * log(p^{(i)}_{target})
\end{equation}
$p^{(i)}_{target}$ is the probability that the element $i$ output by the adaptor is assigned to its correct target value, $\alpha$ is a weight term that corrects for imbalance (computed from the training data) within the binary element-wise ``labels'', and $\gamma$ is a constant controlling how much focus is put on harder-to-classify samples ($gamma=2$ in our experiments).

Beyond enabling easy feature-fusion between adaptors, there are other practical reasons for projecting to target binary class HD vectors at each adaptor. Every class for every adaptor is assigned a unique HD vector. The goal is to map all input instances to the HD vector corresponding to their class label for the given adaptor. To learn this mapping, if each HD vector has a dimensionality of $D$, then a binary classifier is learned for each of the $D$ dimensions. Because each element is assigned either 0 or 1 randomly and all instances from the same class share a target class vector,
this effectively means that the model is learning to predict a random partitioning of the classes into two meta-labels for each element of the HD vector. Because the target HD vectors are mutually orthogonal per class, these $D$ classifiers have low-redundancy. The mapping to an HD vector can be thought of as an ensembling method whereby each adaptor forms a strong multi-class classifier by ensembling many low-redundancy weak binary classifiers. Furthermore, bundling between adaptors is another mathematically-principled form of ensembling. Combining all of these characteristics, the mapping from inputs to a final HD vector embedding ultimately leads to highly discriminative models with high noise tolerance.

\subsubsection{Out-of-Distribution Detection}
To determine if a sample is OOD for a given reconfigurator, the aggregate HD vector is computed and compared against the class prototype HD vectors in the reconfigurator. If the minimum hamming distance to any prototype is larger than a fixed threshold $\tau$, then the sample is predicted to be OOD:
\begin{equation}
ood(x) = \min_{class} \; d_{hamming}(h_{agg}(x), h_{proto}^{class}) > \tau
\label{eq:ood_1}
\end{equation}
This works well if we are only interested in determining if a sample is OOD for a single domain. It is often the case that we need to determine if a sample is OOD over all domains, or we need to identify which set of adaptors/reconfigurators the data should be routed through when knowledge of the true domain is unknown. Such cases may require different thresholds for the reconfigurator, and the OOD scores of each reconfigurator may not be one-to-one comparable (e.g., if it is naturally easier to (over)fit a set of adaptors to one domain vs another, noise characteristics may make distances to the nearest prototype not directly comparable). Thus, we need a calibrated OOD score that is comparable between adaptor sets.

\begin{figure}[t]
{\centering
\includegraphics[width=0.85\linewidth]{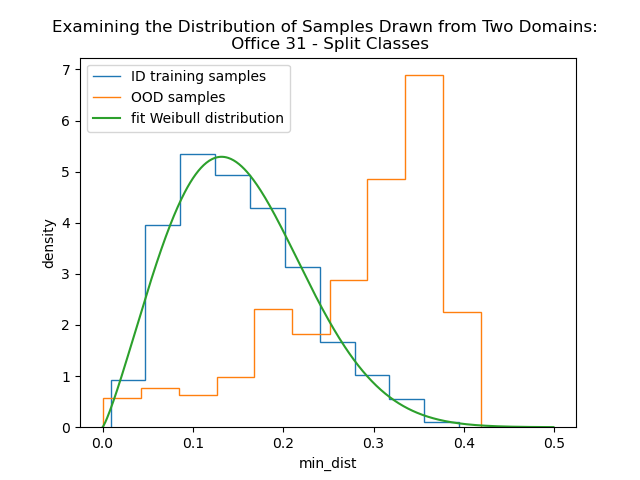}
\caption{Estimating the distribution of ID samples using the Weibull distribution; comparing the densities of ID and OOD data determined by distance to the nearest class prototype}
\label{fig:weibull}
}
\end{figure}

To obtain this calibrated OOD score, we fit a probability distribution over the distances to the nearest prototype for the training set. Building upon work in extreme value theory and open set classification \cite{scheirer2014probability}, we fit a 3-Parameter Weibull distribution \cite{rinne2008weibull} to the ID samples:

\begin{equation}
    PDF_{weib}(x) = 
    \left\{
    \begin{array}{lr}
        \frac{b}{a}\left(\frac{x-c}{a} \right)^{b-1}\exp\left(-\left(\frac{x-c}{a}\right)^{b}\right), & \text{if } x > c\\
        0, & \text{if } x\leq c
    \end{array}
    \right.
    \label{eq:weibull}
\end{equation}
In Eq. \ref{eq:weibull}, $a$ is the scale parameter, $b$ is the shape parameter, and $c$ is the location parameter. These parameters are fit via maximum-likelihood estimation, and generally, the location parameter becomes zero, simplifying to a 2-Parameter Weibull distribution. In Fig. \ref{fig:weibull}, we show a Weibull distribution fit to ID data. Data tends to have a strong right-tail, and thus, we need a more expressive distribution than the Gaussian.

Once the distribution is fit, we can score samples based on how likely the data is to be ID. We use the CDF of the fit Weibull distribution to compute the probability that the distance between a random in-distribution sample and its nearest class prototype is less than the distance between the observed sample and its nearest prototype. We select a hard threshold on the probability $\tau_{\pi}$ to make the final prediction of whether the sample is OOD for each reconfigurator:
\begin{equation}
    ood(x) = CDF_{weib}(\min_{class} \; d_{hamming}(h_{agg}(x), h_{proto}^{class})) > \tau_{\pi}
\end{equation}

\subsection{A Spectral Analysis-Based Approach to Zero-Shot Neural Architecture Search for Adaptation}
\label{sec:zsnas}

Finally, we need to determine how to grow the set of adaptors/reconfigurators as new domains are encountered by employing NAS. Traditional NAS can be incredibly expensive, taking multiple GPU-days to perform. Our goal is for the NAS to be performed on compute-light hardware; thus, we don't have the computational budget to run full NAS. We improve the efficiency of the NAS in two ways: 1) we freeze the feature extracting backbone and only search for the architecture of shallow adaptor layers, significantly reducing the candidate architecture search space and improving model training speed, and 2) we use ZS-NAS where we avoid training the candidate architectures, instead evaluating the candidate model quality via proxy heuristics. We define our NAS as follows:
\begin{itemize}
    \item \textbf{Search Space:} Definition of adaptors/reconfigurators within reasonable parameter bounds specified by a human
    \item \textbf{Search Strategy:} Candidate selection via a global optimizer; we use Bayesian optimization using Gaussian Process Upper Confidence Bounds \cite{snoek2012practical,bayesian} with sequential domain reduction \cite{stander2002robustness} as the acquisition function
    \item \textbf{Performance Estimation Strategy:} Zero-shot proxy heuristic via spectral analysis of the nearest neighbor graph of a random batch of input samples
\end{itemize}

Proxies for ZS-NAS generally aim to maximize one of the following properties: expressivity, trainability, or generalization \cite{chen2021understanding}. These methods generally require computing or approximating gradients over one or more random batches of data. Computing and storing gradients can be computationally- and memory-expensive. We propose a gradient-free proxy heuristic. Furthermore, existing ZS-NAS methods do not assume the use of a frozen backbone network and may not work as expected when used with the EAR framework. Our approach considers the unique properties of the EAR architecture for the resource-constrained use case:
\begin{itemize}
    \item Maximize the expressivity of the representations of each adaptor assuming a fixed pre-trained encoder
    \item Minimize the redundancy of the representations learned across all adaptors
    \item Minimize the number of trainable parameters
\end{itemize}

The first objective of our ZS-NAS is to maximize the expressivity of each adaptor. A random batch of data is passed through each randomly-initialized, untrained adaptor and features are extracted from the layer preceding the final projection to HD space. The Laplacian of the 2-nearest neighbor graph $L$ is computed for this batch of samples in ``random'' feature space. The encoder already extracts useful representations, so we don't want to learn adaptors that degenerate features to a singular point or scramble the features into a uniform space. We hypothesize that an expressive adaptor will result in a rocky feature landscape characterised by small clusters of data points. This can be achieved by maximizing the number of connected components in the nearest neighbor graph. The Laplacian is decomposed into Eigenvalues $\lambda$ and Eigenvectors $\vec{v}$. The number of connected components of the nearest neighbor graph is computed by counting the number of 0-valued Eigenvalues \cite{marsden2013eigenvalues}. We relax the constraint that the score is directly proportional to the number of connected components, and instead, construct a score that uses the number of \textit{loosely} connected components ($\gamma$ controls the strictness of the connected component count; we set $\gamma=3$):
\begin{equation}
    s_{exp}^{ada} = \sum_{i} \max(1-\lambda_{i}^{ada}, 0)^{\gamma}
    \label{eq:expressivity_score}
\end{equation}

The second objective of our approach is to minimize the redundancy of the representations learned across all adaptors. We reuse the Eigenvalues and Eigenvectors from the preceding computation and perform spectral clustering \cite{von2007tutorial} for each adaptor using the approach of Damle et al. \cite{damle2019simple}. The number of clusters is set based on the number of Eigenvalues smaller than 0.1. The redundancy score $s_{red}$ measures cluster overlap using the adjusted mutual information metric \cite{vinh2009information} between all pairs of adaptors.

The final objective is to minimize the number of trainable adaptor/reconfigurator parameters. $s_{par}$ simply counts the number of trainable parameters across all of the adaptor layers.

The final proxy heuristic for our proposed ZS-NAS score is a weighted sum of the three component scores:

\begin{equation}
\begin{split}
    s = & \sum_{all\_adaptors} \left(s_{exp}^{ada} + \beta_{0} * \sum_{all\_adaptors} s_{par}^{ada} \right) + \\
    & \beta_{1} * \sum_{all\_adaptor\_pairs} s_{red}^{ada_{i}, ada_{j}}
    \label{eq:combined_score}
\end{split}
\end{equation}
In our experiments, $\beta_{0}$ is set to $3*10^{-6}$ and $\beta_{1}$ is set to 5.

\section{Results}
We validate our approach on a four of benchmark datasets compared to state-of-the-art benchmark algorithms for OOD detection with DNNs and ZS-NAS.

\subsection{Experimental Setting}
Our experiments are conducted on a machine with two Intel Xeon Gold 6240R CPUs with 24 cores/48 threads each, an NVidia A5000 with 24 GB VRAM, and 500 GB RAM. We implement our models in Python using TensorFlow 2 in compiled mode. Our encoder network is an EfficientNetV2B3 model \cite{tan2021efficientnetv2}. We use the Adam optimizer \cite{kingma2014adam} with a learning rate of 0.001 with no weight decay for learning HD representations, a learning rate of 0.005 with a weight decay parameter of $10^{-5}$ for training the baseline cross-entropy-based NN, and a learning rate of 0.001 and weight decay of $10^{-7}$ for the deep nearest neighbor model trained with the supervised contrastive loss \cite{khosla2020supervised}. For training the NNs, we use a batch size of 128 and train for 40 epochs. For the OOD and ZS-NAS experiments, we run 15 trials with different random seeds per experimental setting and different train/test dataset splits.

\subsection{Datasets}
We evaluate on variants of four benchmark datasets: PACS \cite{li2017deeper}, Office 31 \cite{saenko2010adapting}, Office Home \cite{venkateswara2017deep}, and DomainNet \cite{peng2019moment}. We focus on the use case where data is limited (generally $<100$ training samples per class). We evaluate OOD detection from four perspectives, which requires constructing variants of the aforementioned datasets: 1) both the input modality and output class label sets change (Office Home - Disjoint Domains/Classes), 2) the input modality changes, but the class labels remain the same (PACS), 3) the input modality remains the same, but the class labels change (Office 31 - Split Classes), and 4) the input modality and class labels remain the same, but environmental factors change (Office 31). For our continual learning experiment, we modify the DomainNet dataset by selecting twenty classes from three different modalities, each with 4000 samples, split into six disjoint tasks. Statistics of the datasets appear in Table \ref{tab:datasets}.

\begin{table}[ht!]
\centering
\resizebox{0.49\textwidth}{!}{
\begin{tabular}{|c|c|c|c|c|}
\hline
Dataset & Domain & \#Classes & \#Training Samples & \#Test Samples  \\ \hline \hline
\multicolumn{5}{|c|}{PACS} \\ \hline
PACS & Art/Painting & 7 & 1539 & 509\\ \hline
PACS & Cartoon & 7 & 1760 & 584\\ \hline
PACS & Photo & 7 & 1255 & 415\\ \hline
PACS & Sketch & 7 & 2947 & 982\\ \hline \hline
\multicolumn{5}{|c|}{Office 31} \\ \hline 
Office 31 & Product Images & 31 & 2123 & 693\\ \hline
Office 31 & Webcam & 31 & 608 & 187\\ \hline \hline
\multicolumn{5}{|c|}{Office 31 - Split Classes} \\ \hline 
Office 31 & Product Images - Class Set 1 & 16 & 1082 & 352\\ \hline
Office 31 & Product Images - Class Set 2 & 15 & 1041 & 341\\ \hline \hline
\multicolumn{5}{|c|}{Office Home - Disjoint Domains/Classes} \\ \hline 
Office Home & Art - Class Set 1 & 32 & 908 & 289\\ \hline
Office Home & Clipart - Class Set 1 & 32 & 1688 & 544\\ \hline
Office Home & Real - Class Set 1 & 32 & 1742 & 565\\ \hline
Office Home & Art - Class Set 2 & 33 & 1612 & 521\\ \hline
Office Home & Clipart - Class Set 2 & 33 & 933 & 297\\ \hline
Office Home & Real - Class Set 2 & 33 & 1550 & 500\\ \hline \hline
\multicolumn{5}{|c|}{DomainNet} \\ \hline
DomainNet & Quickdraw - Class Set 1 & 10 & 1000 & 3000\\ \hline
DomainNet & Real - Class Set 1 & 10 & 1000 & 3000\\ \hline
DomainNet & Sketch - Class Set 1 & 10 & 1000 & 3000\\ \hline
DomainNet & Quickdraw - Class Set 2 & 10 & 1000 & 3000\\ \hline
DomainNet & Real - Class Set 2 & 10 & 1000 & 3000\\ \hline
DomainNet & Sketch - Class Set 2 & 10 & 1000 & 3000\\ \hline
\end{tabular}}
\caption{Statistics on the first random split of the datasets used in our experiments}
\label{tab:datasets}
\end{table}

\subsection{Novelty Detection and Classification using Deep HDC}
In the first set of experiments, we take a network with an EfficientNetV2B3 encoder pretrained on Imagenet, and we finetune the entire network on the first domain of interest for image classification and OOD detection. We then apply the network to mixed samples from the test sets of the ID domain and unseen OOD domain. To minimize the effect of the Imagenet pretraining, the first domain always consists of a modality closely aligned to natural images. We consider five training styles: 1) training the encoder plus a 2-layer perceptron using standard cross-entropy, 2) training the encoder plus an embedding layer for deep nearest neighbors using the supervised contrastive loss \cite{sun2022out}, 3) training the encoder plus a single final layer HD adaptor/reconfigurator, 4) training the encoder plus 2-layer perceptron HD adaptors for seven tap points (blocks 1, 4, 7, 12, 19, 31, and ``head'' from the EfficientNetV2B3 model) and a final reconfigurator, and 5) training the encoder, HD adaptors whose tap points (subset of the previous tap points) and architecture are determined using ZS-NAS, and reconfigurator.

To compare with networks not designed explicitly for OOD detection, we train post-hoc OOD detection algorithms on one of i) the features immediately preceding the classification layer, ii) the output softmax probabilities, or iii) using the feature outputs of all 32 layers in the encoder network. For setting (i), we consider off-the-shelf classic OOD and novelty detection algorithms: one-class SVM \cite{manevitz2001one}, empirical and robust covariance estimates \cite{pena2001multivariate}, local outlier factors \cite{breunig2000lof}, and isolation forests \cite{liu2008isolation,liu2012isolation} as implemented in scikit-learn. For setting (ii), we employ the deep baseline for OOD detection proposed by Hendrycks and Gimpel \cite{hendrycks2016baseline}. For setting (iii), we consider two state-of-the-art methods: using layer-wise gram matrices \cite{sastry2020detecting} and hyperdimensional feature fusion \cite{wilson2023hyperdimensional}. We also compare with deep nearest neighbors, which is designed for joint OOD detection and classification.

We consider the following metrics for evaluation: 
\begin{itemize}
    \item Classification accuracy of the ID test samples to measure the discriminability
    \item When the ID and OOD samples share class labels, we also examinw the classification accuracy of the OOD test samples to measure inherent knowledge transfer/robustness to noise from distribution shifts
    \item To measure performance on OOD detection, we compute:
    \begin{itemize}
        \item Area under the receiver-operator curve (AUROC) when ID samples are consider the positive class
        \item Macro-F1-measure of predicting ID vs OOD
        \item True Negative Rate at True Positive Rate 95\% (TNR@TPR95) and 90\% (TNR@TPR90) where the ID data is treated as the positive class
    \end{itemize}
    \item Number of parameters, adaptors, and embedding dimensions to get a sense of the model efficiency and capacity
\end{itemize}
In the cases where a binary decision is necessary, we evaluate against optimal thresholds on the scoring functions. Note that these data sets are relatively challenging for the OOD detection problem; thus, some metrics, even using state-of-the-art algorithms for OOD are low compared to running these algorithms on other OOD detection benchmark datasets. We also apply the Kruskal-Wallis test (non-parametric ANOVA) comparing the performance of the different OOD algorithms within a specific experimental setting followed by the posthoc Dunn test with Bonferroni correction to identify which OOD algorithms differ from the ZS-NAS deep HD-based model in a statistically significant way (p-value of 0.05). In the experimental results tables $\odot$ represents that no statistically significant findings can be determined, $\oplus$ represents the ZS-NAS HD model is significantly better than the comparator, and $\ominus$ represents the ZS-NAS HD model is significantly worse than the comparator.

In Tables \ref{exp1_class:setting1}, \ref{exp1_class:setting2}, \ref{exp1_class:setting3}, and \ref{exp1_class:setting4}, we look at how the classification performance of the proposed approach is effected by various types of domain shift. We notice the following trends:
\begin{itemize}
    \item The learned HD-based approaches outperform the standard and deep nearest neighbor-based approaches w.r.t. ID accuracy by \textasciitilde 2-7\% across all datasets
    \item Na\"{i}vely adding multi-layer perceptron adaptors to all of the candidate tap points can result in slightly improved classification performance, but comes at the cost of a significant increase in the number of model parameters.
    \item The ZS-NAS-determined adaptors generally use significantly smaller architectures while only exhibiting a minor decrease in performance compared to the ``all layers'' setting and outperform the non-HD-based models.
    \item Just learning a single final layer adaptor is a strong baseline, matching performance with the ZS-NAS model in terms of classification accuracy in most cases.
    \item Interestingly, when applying the model trained on one domain to a different domain with the same class set, the learned HD-based models significantly outperform the non-HD-based models in terms of accuracy, suggesting that the HD representation is more robust to noise caused by distribution shift. This is especially apparent for perceptually-similar domains, e.g., transferring from natural images to photorealistic paintings. The ``all layers'' model generalizes worse than the ZS-NAS model.
\end{itemize}

In Tables \ref{exp1_ood:setting1}, \ref{exp1_ood:setting2}, \ref{exp1_ood:setting3}, and \ref{exp1_ood:setting4}, we look at how OOD detection performance is effected by various types of domain shift. We notice the following trends:
\begin{itemize}
\item Regardless of setting, the learned HD-based models match or exceed the performance of the other models.
\item The ``all layers'' HD model occasionally underperforms on the OOD sample detection task compared to the ``final layer'' and ZS-NAS HD models (e.g., in the PACS photo-to-sketch transfer task). This is especially apparent when looking at the TNR@TPR metrics.
\end{itemize}

\begin{table}[h!]
\centering
\resizebox{0.49\textwidth}{!}{
\begin{tabular}{|c|c|c|c|c|}
\hline
Method & Accuracy\_ID & \#Parameters & \#Adaptors & Embed\_Dim \\ \hline 
\multicolumn{5}{|c|}{office\_home\_disjoint - real\_disjoint : art\_disjoint} \\ \hline 
standard\_training & $0.68\pm0.09\odot$ &  $14559270.0$ &$1.0$ &$128.0$ \\ \hline 
scl & $0.64\pm0.07\oplus$ & $14559270.0$ &$1.0$ &$128.0$ \\ \hline 
learned\_hd\_final\_layer & $0.73\pm0.05\odot$ & $14459269.0$ &$1.0$ &$63.0$ \\ \hline 
learned\_hd\_all\_layers & $0.74\pm0.05\odot$ & $124864951.0$ &$7.0$ &$255.0$ \\ \hline 
learned\_hd\_zs\_nas & $0.73\pm0.06\odot$ & $13414837.8$ &$6.07$ &$255.0$ \\ \hline 
\multicolumn{5}{|c|}{office\_home\_disjoint - real\_disjoint : clipart\_disjoint} \\ \hline 
standard\_training & $0.68\pm0.09\odot$ &  $14559270.0$ &$1.0$ &$128.0$ \\ \hline 
scl & $0.64\pm0.07\oplus$ & $14559270.0$ &$1.0$ &$128.0$ \\ \hline 
learned\_hd\_final\_layer & $0.73\pm0.05\odot$ & $14459269.0$ &$1.0$ &$63.0$ \\ \hline 
learned\_hd\_all\_layers & $0.74\pm0.05\odot$ & $124864951.0$ &$7.0$ &$255.0$ \\ \hline 
learned\_hd\_zs\_nas & $0.73\pm0.06\odot$ & $13414837.8$ &$6.07$ &$255.0$ \\ \hline 
\end{tabular}}
\caption{Comparing classification performance on the Office Home - Disjoint Domains/Classes datasets where input modality and output class label sets change across domains}
\label{exp1_class:setting1}
\end{table}

\begin{table}[h!]
\centering
\resizebox{0.49\textwidth}{!}{
\begin{tabular}{|c|c|c|c|c|}
\hline
Method & AUROC & macro-F1 & TNR@TPR95 & TNR@TPR90 \\ \hline 
\multicolumn{5}{|c|}{office\_home\_disjoint - real\_disjoint : art\_disjoint} \\ \hline 
oc-svm & $0.78\pm0.04\odot$ & $0.45\pm0.07\oplus$ & $0.19\pm0.06\odot$ & $0.35\pm0.09\odot$ \\ \hline 
empirical\_cov & $0.76\pm0.07\odot$ & $0.61\pm0.07\oplus$ & $0.19\pm0.08\odot$ & $0.33\pm0.1\odot$ \\ \hline 
robust\_cov & $0.73\pm0.06\odot$ & $0.61\pm0.08\oplus$ & $0.15\pm0.07\odot$ & $0.28\pm0.1\odot$ \\ \hline 
lof & $0.73\pm0.06\odot$ & $0.58\pm0.06\oplus$ & $0.12\pm0.05\oplus$ & $0.26\pm0.09\odot$ \\ \hline 
isof & $0.76\pm0.07\odot$ & $0.64\pm0.04\oplus$ & $0.2\pm0.08\odot$ & $0.34\pm0.1\odot$ \\ \hline 
deep\_baseline & $0.73\pm0.06\odot$ & $0.67\pm0.04\odot$ & $0.18\pm0.05\odot$ & $0.3\pm0.08\odot$ \\ \hline 
gram & $0.7\pm0.06\oplus$ & $0.65\pm0.05\odot$ & $0.12\pm0.04\oplus$ & $0.25\pm0.08\odot$ \\ \hline 
hd\_fusion & $0.67\pm0.07\oplus$ & $0.62\pm0.05\oplus$ & $0.09\pm0.04\oplus$ & $0.19\pm0.07\oplus$ \\ \hline 
scl & $0.69\pm0.06\oplus$ & $0.63\pm0.05\oplus$ & $0.1\pm0.05\oplus$ & $0.2\pm0.06\oplus$ \\ \hline 
learned\_hd\_final\_layer & $0.79\pm0.04\odot$ & $0.72\pm0.03\odot$ & $0.18\pm0.06\odot$ & $0.35\pm0.08\odot$ \\ \hline 
learned\_hd\_all\_layers & $0.81\pm0.03\odot$ & $0.73\pm0.03\odot$ & $0.21\pm0.05\odot$ & $0.37\pm0.06\odot$ \\ \hline 
learned\_hd\_zs\_nas & $0.8\pm0.03\odot$ & $0.72\pm0.03\odot$ & $0.22\pm0.07\odot$ & $0.38\pm0.09\odot$ \\ \hline 
\multicolumn{5}{|c|}{office\_home\_disjoint - real\_disjoint : clipart\_disjoint} \\ \hline 
oc-svm & $0.75\pm0.04\odot$ & $0.51\pm0.06\oplus$ & $0.17\pm0.04\odot$ & $0.31\pm0.06\odot$ \\ \hline 
empirical\_cov & $0.73\pm0.06\odot$ & $0.61\pm0.09\oplus$ & $0.14\pm0.06\odot$ & $0.27\pm0.09\odot$ \\ \hline 
robust\_cov & $0.72\pm0.06\odot$ & $0.57\pm0.13\oplus$ & $0.12\pm0.08\odot$ & $0.24\pm0.1\odot$ \\ \hline 
lof & $0.71\pm0.04\odot$ & $0.6\pm0.07\oplus$ & $0.11\pm0.05\oplus$ & $0.23\pm0.07\odot$ \\ \hline 
isof & $0.73\pm0.06\odot$ & $0.65\pm0.05\odot$ & $0.18\pm0.06\odot$ & $0.31\pm0.09\odot$ \\ \hline 
deep\_baseline & $0.72\pm0.05\odot$ & $0.68\pm0.04\odot$ & $0.16\pm0.04\odot$ & $0.29\pm0.06\odot$ \\ \hline 
gram & $0.77\pm0.04\odot$ & $0.71\pm0.03\odot$ & $0.19\pm0.06\odot$ & $0.34\pm0.08\odot$ \\ \hline 
hd\_fusion & $0.83\pm0.05\odot$ & $0.76\pm0.04\odot$ & $0.33\pm0.12\odot$ & $0.49\pm0.14\odot$ \\ \hline 
scl & $0.65\pm0.07\oplus$ & $0.61\pm0.05\oplus$ & $0.12\pm0.04\oplus$ & $0.19\pm0.06\oplus$ \\ \hline 
learned\_hd\_final\_layer & $0.77\pm0.03\odot$ & $0.72\pm0.02\odot$ & $0.17\pm0.04\odot$ & $0.32\pm0.05\odot$ \\ \hline 
learned\_hd\_all\_layers & $0.77\pm0.04\odot$ & $0.71\pm0.04\odot$ & $0.17\pm0.05\odot$ & $0.31\pm0.06\odot$ \\ \hline 
learned\_hd\_zs\_nas & $0.77\pm0.04\odot$ & $0.72\pm0.04\odot$ & $0.2\pm0.06\odot$ & $0.34\pm0.09\odot$ \\ \hline 
\end{tabular}}
\caption{Comparing OOD detection performance on the Office Home - Disjoint Domains/Classes datasets where input modality and output class label sets change across domains}
\label{exp1_ood:setting1}
\end{table}

\begin{table}[h!]
\centering
\resizebox{0.49\textwidth}{!}{
\begin{tabular}{|c|c|c|c|c|c|}
\hline
Method & Accuracy\_ID & Accuracy\_OOD & \#Parameters & \#Adaptors & Embed\_Dim \\ \hline 
\multicolumn{6}{|c|}{pacs - photo : cartoon} \\ \hline 
standard\_training & $0.89\pm0.02\oplus$ & $0.19\pm0.05\oplus$ & $14556045.0$ &$1.0$ &$128.0$ \\ \hline 
scl & $0.89\pm0.02\oplus$ & $0.18\pm0.04\oplus$ & $14556045.0$ &$1.0$ &$128.0$ \\ \hline 
learned\_hd\_final\_layer & $0.96\pm0.01\odot$ & $0.26\pm0.05\odot$ & $14459269.0$ &$1.0$ &$63.0$ \\ \hline 
learned\_hd\_all\_layers & $0.96\pm0.01\odot$ & $0.15\pm0.02\oplus$ & $41575351.0$ &$7.0$ &$63.0$ \\ \hline 
learned\_hd\_zs\_nas & $0.95\pm0.02\odot$ & $0.26\pm0.06\odot$ & $13400336.73$ &$6.07$ &$63.0$ \\ \hline 
\multicolumn{6}{|c|}{pacs - photo : painting} \\ \hline 
standard\_training & $0.89\pm0.02\oplus$ & $0.41\pm0.04\oplus$ & $14556045.0$ &$1.0$ &$128.0$ \\ \hline 
scl & $0.89\pm0.02\oplus$ & $0.4\pm0.03\oplus$ & $14556045.0$ &$1.0$ &$128.0$ \\ \hline 
learned\_hd\_final\_layer & $0.96\pm0.01\odot$ & $0.55\pm0.05\odot$ & $14459269.0$ &$1.0$ &$63.0$ \\ \hline 
learned\_hd\_all\_layers & $0.96\pm0.01\odot$ & $0.5\pm0.03\odot$ & $41575351.0$ &$7.0$ &$63.0$ \\ \hline 
learned\_hd\_zs\_nas & $0.95\pm0.02\odot$ & $0.52\pm0.07\odot$ & $13400336.73$ &$6.07$ &$63.0$ \\ \hline 
\multicolumn{6}{|c|}{pacs - photo : sketch} \\ \hline 
standard\_training & $0.89\pm0.02\oplus$ & $0.22\pm0.08\odot$ & $14556045.0$ &$1.0$ &$128.0$ \\ \hline 
scl & $0.89\pm0.02\oplus$ & $0.21\pm0.06\odot$ & $14556045.0$ &$1.0$ &$128.0$ \\ \hline 
learned\_hd\_final\_layer & $0.96\pm0.01\odot$ & $0.27\pm0.08\odot$ & $14459269.0$ &$1.0$ &$63.0$ \\ \hline 
learned\_hd\_all\_layers & $0.96\pm0.01\odot$ & $0.17\pm0.01\oplus$ & $41575351.0$ &$7.0$ &$63.0$ \\ \hline 
learned\_hd\_zs\_nas & $0.95\pm0.02\odot$ & $0.29\pm0.09\odot$ & $13400336.73$ &$6.07$ &$63.0$ \\ \hline 
\end{tabular}}
\caption{Comparing classification performance on the PACS datasets where input modalities change, but the output class label remains the same across domains}
\label{exp1_class:setting2}
\end{table}

\begin{table}[h!]
\centering
\resizebox{0.49\textwidth}{!}{
\begin{tabular}{|c|c|c|c|c|}
\hline
Method & AUROC & macro-F1 & TNR@TPR95 & TNR@TPR90 \\ \hline 
\multicolumn{5}{|c|}{pacs - photo : cartoon} \\ \hline 
oc-svm & $0.67\pm0.07\odot$ & $0.58\pm0.08\odot$ & $0.11\pm0.06\odot$ & $0.2\pm0.11\odot$ \\ \hline 
empirical\_cov & $0.64\pm0.09\odot$ & $0.43\pm0.09\oplus$ & $0.04\pm0.03\oplus$ & $0.12\pm0.08\oplus$ \\ \hline 
robust\_cov & $0.66\pm0.11\odot$ & $0.54\pm0.16\odot$ & $0.1\pm0.11\oplus$ & $0.17\pm0.17\oplus$ \\ \hline 
lof & $0.69\pm0.07\odot$ & $0.55\pm0.08\oplus$ & $0.15\pm0.06\odot$ & $0.25\pm0.08\odot$ \\ \hline 
isof & $0.68\pm0.09\odot$ & $0.58\pm0.15\odot$ & $0.07\pm0.04\oplus$ & $0.16\pm0.09\oplus$ \\ \hline 
deep\_baseline & $0.57\pm0.16\oplus$ & $0.6\pm0.12\odot$ & $0.07\pm0.04\oplus$ & $0.14\pm0.07\oplus$ \\ \hline 
gram & $0.6\pm0.13\odot$ & $0.59\pm0.1\odot$ & $0.13\pm0.1\odot$ & $0.23\pm0.14\odot$ \\ \hline 
hd\_fusion & $0.82\pm0.04\odot$ & $0.78\pm0.04\odot$ & $0.23\pm0.15\odot$ & $0.39\pm0.18\odot$ \\ \hline 
scl & $0.69\pm0.08\odot$ & $0.7\pm0.06\odot$ & $0.06\pm0.03\oplus$ & $0.15\pm0.06\oplus$ \\ \hline 
learned\_hd\_final\_layer & $0.8\pm0.07\odot$ & $0.76\pm0.07\odot$ & $0.29\pm0.14\odot$ & $0.48\pm0.17\odot$ \\ \hline 
learned\_hd\_all\_layers & $0.73\pm0.04\odot$ & $0.72\pm0.02\odot$ & $0.08\pm0.05\oplus$ & $0.17\pm0.09\oplus$ \\ \hline 
learned\_hd\_zs\_nas & $0.74\pm0.11\odot$ & $0.71\pm0.09\odot$ & $0.25\pm0.12\odot$ & $0.4\pm0.16\odot$ \\ \hline 
\multicolumn{5}{|c|}{pacs - photo : painting} \\ \hline 
oc-svm & $0.75\pm0.04\oplus$ & $0.72\pm0.03\oplus$ & $0.22\pm0.06\oplus$ & $0.39\pm0.07\oplus$ \\ \hline 
empirical\_cov & $0.79\pm0.03\odot$ & $0.63\pm0.09\oplus$ & $0.18\pm0.08\oplus$ & $0.33\pm0.09\oplus$ \\ \hline 
robust\_cov & $0.79\pm0.03\odot$ & $0.7\pm0.09\oplus$ & $0.19\pm0.08\oplus$ & $0.33\pm0.09\oplus$ \\ \hline 
lof & $0.77\pm0.04\oplus$ & $0.64\pm0.08\oplus$ & $0.2\pm0.05\oplus$ & $0.34\pm0.05\oplus$ \\ \hline 
isof & $0.78\pm0.04\odot$ & $0.69\pm0.05\oplus$ & $0.2\pm0.07\oplus$ & $0.38\pm0.08\oplus$ \\ \hline 
deep\_baseline & $0.7\pm0.06\oplus$ & $0.69\pm0.04\oplus$ & $0.18\pm0.06\oplus$ & $0.33\pm0.07\oplus$ \\ \hline 
gram & $0.66\pm0.07\oplus$ & $0.63\pm0.07\oplus$ & $0.16\pm0.05\oplus$ & $0.28\pm0.08\oplus$ \\ \hline 
hd\_fusion & $0.77\pm0.03\oplus$ & $0.72\pm0.03\oplus$ & $0.14\pm0.06\oplus$ & $0.29\pm0.09\oplus$ \\ \hline 
scl & $0.83\pm0.03\odot$ & $0.77\pm0.03\odot$ & $0.24\pm0.07\odot$ & $0.42\pm0.08\odot$ \\ \hline 
learned\_hd\_final\_layer & $0.83\pm0.02\odot$ & $0.78\pm0.02\odot$ & $0.38\pm0.06\odot$ & $0.57\pm0.06\odot$ \\ \hline 
learned\_hd\_all\_layers & $0.87\pm0.01\odot$ & $0.8\pm0.02\odot$ & $0.34\pm0.05\odot$ & $0.55\pm0.06\odot$ \\ \hline 
learned\_hd\_zs\_nas & $0.84\pm0.02\odot$ & $0.79\pm0.02\odot$ & $0.38\pm0.08\odot$ & $0.58\pm0.07\odot$ \\ \hline 
\multicolumn{5}{|c|}{pacs - photo : sketch} \\ \hline 
oc-svm & $0.63\pm0.17\odot$ & $0.51\pm0.19\oplus$ & $0.14\pm0.18\odot$ & $0.23\pm0.24\odot$ \\ \hline 
empirical\_cov & $0.66\pm0.17\odot$ & $0.44\pm0.18\oplus$ & $0.1\pm0.14\oplus$ & $0.22\pm0.23\odot$ \\ \hline 
robust\_cov & $0.72\pm0.16\odot$ & $0.62\pm0.17\odot$ & $0.17\pm0.19\odot$ & $0.27\pm0.23\odot$ \\ \hline 
lof & $0.65\pm0.19\odot$ & $0.48\pm0.18\oplus$ & $0.15\pm0.15\odot$ & $0.26\pm0.23\odot$ \\ \hline 
isof & $0.75\pm0.12\odot$ & $0.64\pm0.18\odot$ & $0.17\pm0.21\odot$ & $0.28\pm0.28\odot$ \\ \hline 
deep\_baseline & $0.54\pm0.21\oplus$ & $0.61\pm0.14\odot$ & $0.06\pm0.06\oplus$ & $0.11\pm0.11\oplus$ \\ \hline 
gram & $0.82\pm0.12\odot$ & $0.8\pm0.09\odot$ & $0.28\pm0.25\odot$ & $0.46\pm0.3\odot$ \\ \hline 
hd\_fusion & $0.94\pm0.05\odot$ & $0.92\pm0.04\odot$ & $0.66\pm0.32\odot$ & $0.76\pm0.29\odot$ \\ \hline 
scl & $0.81\pm0.1\odot$ & $0.79\pm0.08\odot$ & $0.27\pm0.22\odot$ & $0.4\pm0.27\odot$ \\ \hline 
learned\_hd\_final\_layer & $0.84\pm0.11\odot$ & $0.79\pm0.1\odot$ & $0.42\pm0.26\odot$ & $0.59\pm0.27\odot$ \\ \hline 
learned\_hd\_all\_layers & $0.69\pm0.08\odot$ & $0.74\pm0.03\odot$ & $0.05\pm0.07\oplus$ & $0.11\pm0.13\oplus$ \\ \hline 
learned\_hd\_zs\_nas & $0.81\pm0.21\odot$ & $0.78\pm0.12\odot$ & $0.4\pm0.26\odot$ & $0.55\pm0.28\odot$ \\ \hline 
\end{tabular}}
\caption{Comparing the OOD detection performance on the PACS datasets where input modalities change, but the output class label remains the same across domains}
\label{exp1_ood:setting2}
\end{table}

\begin{table}[h!]
\centering
\resizebox{0.49\textwidth}{!}{
\begin{tabular}{|c|c|c|c|c|}
\hline
Method & Accuracy\_ID & \#Parameters & \#Adaptors & Embed\_Dim \\ \hline 
\multicolumn{5}{|c|}{office\_31\_split\_classes - amazon\_split\_1 : amazon\_split\_2} \\ \hline 
standard\_training & $0.79\pm0.05\oplus$ & $14557206.0$ &$1.0$ &$128.0$ \\ \hline 
scl & $0.8\pm0.04\odot$ & $14557206.0$ &$1.0$ &$128.0$ \\ \hline 
learned\_hd\_final\_layer & $0.84\pm0.03\odot$ & $14459269.0$ &$1.0$ &$63.0$ \\ \hline 
learned\_hd\_all\_layers & $0.85\pm0.03\odot$ & $69281207.0$ &$7.0$ &$127.0$ \\ \hline 
learned\_hd\_zs\_nas & $0.84\pm0.04\odot$ & $13578750.27$ &$6.13$ &$127.0$ \\ \hline 
\end{tabular}}
\caption{Comparing classification performance on the Office 31 - Split Classes datasets where input modalities remain the same, but output class labels change}
\label{exp1_class:setting3}
\end{table}

\begin{table}[h!]
\centering
\resizebox{0.49\textwidth}{!}{
\begin{tabular}{|c|c|c|c|c|}
\hline
Method & AUROC & macro-F1 & TNR@TPR95 & TNR@TPR90 \\ \hline 
\multicolumn{5}{|c|}{office\_31\_split\_classes - amazon\_split\_1 : amazon\_split\_2} \\ \hline 
oc-svm & $0.81\pm0.03\odot$ & $0.69\pm0.03\oplus$ & $0.21\pm0.06\odot$ & $0.38\pm0.07\odot$ \\ \hline 
empirical\_cov & $0.79\pm0.04\odot$ & $0.67\pm0.06\oplus$ & $0.18\pm0.05\odot$ & $0.33\pm0.08\odot$ \\ \hline 
robust\_cov & $0.79\pm0.04\odot$ & $0.7\pm0.04\oplus$ & $0.19\pm0.06\odot$ & $0.33\pm0.08\odot$ \\ \hline 
lof & $0.79\pm0.04\odot$ & $0.71\pm0.06\odot$ & $0.18\pm0.07\odot$ & $0.32\pm0.08\odot$ \\ \hline 
isof & $0.77\pm0.05\odot$ & $0.7\pm0.03\oplus$ & $0.18\pm0.05\odot$ & $0.33\pm0.08\odot$ \\ \hline 
deep\_baseline & $0.75\pm0.05\oplus$ & $0.72\pm0.04\odot$ & $0.2\pm0.05\odot$ & $0.35\pm0.09\odot$ \\ \hline 
gram & $0.73\pm0.04\oplus$ & $0.69\pm0.03\oplus$ & $0.16\pm0.04\odot$ & $0.29\pm0.06\odot$ \\ \hline 
hd\_fusion & $0.74\pm0.05\oplus$ & $0.71\pm0.04\oplus$ & $0.09\pm0.05\oplus$ & $0.19\pm0.08\oplus$ \\ \hline 
scl & $0.78\pm0.04\odot$ & $0.74\pm0.03\odot$ & $0.13\pm0.07\odot$ & $0.24\pm0.1\oplus$ \\ \hline 
learned\_hd\_final\_layer & $0.83\pm0.03\odot$ & $0.78\pm0.03\odot$ & $0.23\pm0.06\odot$ & $0.43\pm0.09\odot$ \\ \hline 
learned\_hd\_all\_layers & $0.85\pm0.02\odot$ & $0.8\pm0.02\odot$ & $0.25\pm0.06\odot$ & $0.43\pm0.07\odot$ \\ \hline 
learned\_hd\_zs\_nas & $0.83\pm0.02\odot$ & $0.77\pm0.02\odot$ & $0.2\pm0.05\odot$ & $0.4\pm0.06\odot$ \\ \hline 
\end{tabular}}
\caption{Comparing the OOD detection performance on the Office 31 - Split Classes datasets where input modalities remain the same, but output class labels change}
\label{exp1_ood:setting3}
\end{table}

\begin{table}[h!]
\centering
\resizebox{0.49\textwidth}{!}{
\begin{tabular}{|c|c|c|c|c|c|}
\hline
Method & Accuracy\_ID & Accuracy\_OOD & \#Parameters & \#Adaptors & Embed\_Dim \\ \hline 
\multicolumn{6}{|c|}{office\_31 - amazon : webcam} \\ \hline 
standard\_training & $0.79\pm0.03\odot$ & $0.35\pm0.05\oplus$ & $14559141.0$ &$1.0$ &$128.0$ \\ \hline 
scl & $0.79\pm0.02\odot$ & $0.12\pm0.04\oplus$ & $14559141.0$ &$1.0$ &$128.0$ \\ \hline 
learned\_hd\_final\_layer & $0.8\pm0.02\odot$ & $0.41\pm0.05\odot$ & $14459269.0$ &$1.0$ &$63.0$ \\ \hline 
learned\_hd\_all\_layers & $0.81\pm0.02\odot$ & $0.37\pm0.04\odot$ & $124864951.0$ &$7.0$ &$255.0$ \\ \hline 
learned\_hd\_zs\_nas & $0.8\pm0.02\odot$ & $0.43\pm0.04\odot$ & $13612005.33$ &$6.13$ &$255.0$ \\ \hline 
\end{tabular}}
\caption{Comparing classification performance on the Office 31 datasets where input modalities and class labels remain the same across domains, but the environmental factors (resolution of the imagery) change between domains}
\label{exp1_class:setting4}
\end{table}

\begin{table}[h!]
\centering
\resizebox{0.49\textwidth}{!}{
\begin{tabular}{|c|c|c|c|c|}
\hline
Method & AUROC & macro-F1 & TNR@TPR95 & TNR@TPR90 \\ \hline 
\multicolumn{5}{|c|}{office\_31 - amazon : webcam} \\ \hline 
oc-svm & $0.82\pm0.02\odot$ & $0.53\pm0.02\oplus$ & $0.23\pm0.05\odot$ & $0.41\pm0.04\odot$ \\ \hline 
empirical\_cov & $0.81\pm0.03\odot$ & $0.66\pm0.02\odot$ & $0.22\pm0.06\odot$ & $0.39\pm0.1\odot$ \\ \hline 
robust\_cov & $0.78\pm0.04\odot$ & $0.58\pm0.08\oplus$ & $0.16\pm0.1\odot$ & $0.32\pm0.14\odot$ \\ \hline 
lof & $0.81\pm0.03\odot$ & $0.62\pm0.03\oplus$ & $0.17\pm0.04\odot$ & $0.34\pm0.06\odot$ \\ \hline 
isof & $0.8\pm0.04\odot$ & $0.61\pm0.03\oplus$ & $0.26\pm0.07\odot$ & $0.44\pm0.08\ominus$ \\ \hline 
deep\_baseline & $0.77\pm0.05\odot$ & $0.68\pm0.04\odot$ & $0.19\pm0.06\odot$ & $0.34\pm0.08\odot$ \\ \hline 
gram & $0.74\pm0.03\odot$ & $0.65\pm0.03\odot$ & $0.17\pm0.04\odot$ & $0.31\pm0.05\odot$ \\ \hline 
hd\_fusion & $0.73\pm0.04\odot$ & $0.62\pm0.03\odot$ & $0.1\pm0.03\odot$ & $0.2\pm0.04\odot$ \\ \hline 
scl & $0.85\pm0.01\ominus$ & $0.73\pm0.02\odot$ & $0.25\pm0.05\odot$ & $0.43\pm0.05\ominus$ \\ \hline 
learned\_hd\_final\_layer & $0.78\pm0.03\odot$ & $0.68\pm0.02\odot$ & $0.18\pm0.05\odot$ & $0.33\pm0.05\odot$ \\ \hline 
learned\_hd\_all\_layers & $0.85\pm0.01\ominus$ & $0.72\pm0.02\odot$ & $0.24\pm0.04\odot$ & $0.44\pm0.05\ominus$ \\ \hline 
learned\_hd\_zs\_nas & $0.79\pm0.02\odot$ & $0.69\pm0.02\odot$ & $0.19\pm0.03\odot$ & $0.33\pm0.05\odot$ \\ \hline 
\end{tabular}
}
\caption{Comparing the OOD detection performance on the Office 31 datasets where the input modalities and class labels remain the same across domains, but the environmental factors (resolution of the imagery) change between domains}
\label{exp1_ood:setting4}
\end{table}

\subsection{Comparison of Different NAS Methods}

In the second set of experiments, we compare the proposed ZS-NAS method (``spectral'') to existing NAS methods. We finetune the encoder and a learn a set of adaptors on the first domain, and then freeze the encoder and learn a set of adaptors on the second domain. We compare against adding a final layer adaptor and finetuning the entire network on the second task, adding a final layer adaptor without finetuning the encoder, randomly selecting the location and architectures of the adaptor set, using few-shot NAS (with a single training epoch per candidate evaluation), and comparing to other zero-shot proxy heuristics: computing the $\ell_{2}$-norm of the gradients (grad\_norm) of the candidate architectures, synflow (designed to maximize the sparsity of the candidate architectures) \cite{tanaka2020pruning}, and the $\phi$-score \cite{lin2021zen} and Jacobian-covariance (jacob\_cov) scores \cite{mellor2021neural}, which try to maximize network expressivity. In all cases, we add $s_{par}$ (scaled by trial-and-error) to the base heuristic to constrain the memory footprint of the adaptor set. 

We evaluate the approaches based on the accuracy of the adaptor set on the new domain test data and in terms of computational and memory efficiency as determined by the number of trainable parameters, time to perform the NAS, number of adaptors, and size of embedding. Again we use the Kruskal-Wallis ANOVA test with posthoc Dunn test with Bonferroni correction (p-value of 0.05). Results appear in Table \ref{tab:zsnas}. We observe:
\begin{itemize}
    \item The spectral method achieves accuracy equal or higher to the next best ZS-NAS approach in all settings.
    \item In most cases, the spectral method is as good or better than finetuning the entire network in terms of accuracy.
    \item Once again, learning a final layer adaptor is a surprisingly strong baseline.
    \item The spectral method is 2-7x faster for heuristic evaluation compared to other ZS-NAS methods (note that these methods were re-implemented by the authors and may not necessarily be fully optimized).
    \item The size of the adaptor sets identified by the proposed method are generally within the same order of magnitude as the other approaches, but the spectral method seems more flexible to growing and shrinking as needed. For example, on the Office 31 dataset, the spectral method finds a model roughly 2-3 times the size of the other methods but achieves a 13\% increase in accuracy compared to the next-best ZS-NAS method.
\end{itemize}

\begin{table*}
\centering
\resizebox{0.75\textwidth}{!}{
\begin{tabular}{|c|c|c|c|c|c|}
\hline
Metric & Accuracy\_New & \#Trainable\_Parameters & Search\_Time (s) & \#Adaptors & Embed\_Dim \\ \hline 
\hline
\multicolumn{6}{|c|}{office\_31 - amazon : webcam} \\ \hline 
finetune\_all\_layers & $0.9\pm0.08\odot$ & $14459269.0\pm0.0\odot$ & $0.0\pm0.0\odot$ & $1.0$ &$63.0$ \\ \hline 
finetune\_final\_layer & $0.94\pm0.02\odot$ & $100863.0\pm0.0\ominus$ & $0.0\pm0.0\odot$ & $1.0$ &$63.0$ \\ \hline 
random & $0.81\pm0.36\odot$ & $3878524.67\pm4317844.29\odot$ & $0.0\pm0.0\odot$ & $5.87$ &$246.47$ \\ \hline 
few\_shot & $0.84\pm0.02\oplus$ & $389977.6\pm26599.17\odot$ & $3375.02\pm420.16\oplus$ & $6.67$ &$255.0$ \\ \hline 
grad\_norm & $0.8\pm0.03\oplus$ & $370988.87\pm27955.23\ominus$ & $1000.45\pm14.21\odot$ & $6.33$ &$255.0$ \\ \hline 
synflow & $0.82\pm0.03\oplus$ & $377331.13\pm28226.82\odot$ & $1920.09\pm11.64\oplus$ & $6.47$ &$255.0$ \\ \hline 
$\phi$-score & $0.83\pm0.03\oplus$ & $369056.0\pm28895.17\ominus$ & $1898.06\pm23.3\oplus$ & $6.4$ &$255.0$ \\ \hline 
jacob\_cov & $0.79\pm0.02\oplus$ & $348685.33\pm21601.64\ominus$ & $893.43\pm39.43\odot$ & $5.87$ &$255.0$ \\ \hline 
spectral & $0.97\pm0.01\odot$ & $889030.87\pm158014.84\odot$ & $359.72\pm15.61\odot$ & $5.93$ &$255.0$ \\ \hline 
\hline
\multicolumn{6}{|c|}{office\_31\_split\_classes - amazon\_split\_1 : amazon\_split\_2} \\ \hline 
finetune\_all\_layers & $0.83\pm0.05\oplus$ & $14459269.0\pm0.0\odot$ & $0.0\pm0.0\odot$ & $1.0$ &$63.0$ \\ \hline 
finetune\_final\_layer & $0.89\pm0.02\odot$ & $100863.0\pm0.0\ominus$ & $0.0\pm0.0\odot$ & $1.0$ &$63.0$ \\ \hline 
random & $0.79\pm0.3\odot$ & $3326234.93\pm2734746.09\odot$ & $0.0\pm0.0\odot$ & $6.0$ &$127.0$ \\ \hline 
few\_shot & $0.89\pm0.02\odot$ & $358857.13\pm26543.18\odot$ & $3245.72\pm455.65\oplus$ & $6.6$ &$127.0$ \\ \hline 
grad\_norm & $0.89\pm0.02\odot$ & $340185.73\pm24446.33\odot$ & $995.79\pm47.31\odot$ & $6.27$ &$127.0$ \\ \hline 
synflow & $0.89\pm0.02\odot$ & $352025.47\pm29209.71\odot$ & $1866.85\pm18.85\oplus$ & $6.53$ &$127.0$ \\ \hline 
$\phi$-score & $0.88\pm0.02\odot$ & $345580.73\pm27199.91\odot$ & $1850.41\pm22.53\oplus$ & $6.47$ &$127.0$ \\ \hline 
jacob\_cov & $0.9\pm0.02\odot$ & $316406.8\pm32323.29\ominus$ & $773.48\pm43.44\odot$ & $5.73$ &$127.0$ \\ \hline 
spectral & $0.9\pm0.02\odot$ & $758325.73\pm203864.54\odot$ & $297.61\pm8.54\odot$ & $6.0$ &$127.0$ \\ \hline 
\hline
\multicolumn{6}{|c|}{office\_home\_disjoint - real\_disjoint : clipart\_disjoint} \\ \hline 
finetune\_all\_layers & $0.77\pm0.04\odot$ & $14459269.0\pm0.0\oplus$ & $0.0\pm0.0\odot$ & $1.0$ &$63.0$ \\ \hline 
finetune\_final\_layer & $0.78\pm0.03\odot$ & $100863.0\pm0.0\ominus$ & $0.0\pm0.0\odot$ & $1.0$ &$63.0$ \\ \hline 
random & $0.49\pm0.4\odot$ & $5836242.73\pm5743305.71\odot$ & $0.0\pm0.0\odot$ & $5.67$ &$255.0$ \\ \hline 
few\_shot & $0.73\pm0.05\odot$ & $386918.93\pm25202.76\odot$ & $4528.15\pm422.91\oplus$ & $6.67$ &$255.0$ \\ \hline 
grad\_norm & $0.67\pm0.04\odot$ & $351930.0\pm0.0\odot$ & $998.3\pm10.84\odot$ & $6.0$ &$255.0$ \\ \hline 
synflow & $0.7\pm0.05\odot$ & $365740.93\pm25850.9\odot$ & $1925.56\pm17.63\oplus$ & $6.27$ &$255.0$ \\ \hline 
$\phi$-score & $0.7\pm0.19\odot$ & $485802.47\pm382769.61\odot$ & $1881.11\pm52.96\oplus$ & $6.6$ &$255.0$ \\ \hline 
jacob\_cov & $0.69\pm0.05\odot$ & $346080.53\pm29502.77\odot$ & $828.48\pm45.21\odot$ & $5.87$ &$255.0$ \\ \hline 
spectral & $0.73\pm0.06\odot$ & $423169.93\pm126290.1\odot$ & $349.68\pm11.81\odot$ & $6.33$ &$255.0$ \\ \hline 
\hline
\multicolumn{6}{|c|}{pacs - photo : sketch} \\ \hline 
finetune\_all\_layers & $0.89\pm0.04\odot$ & $14459269.0\pm0.0\oplus$ & $0.0\pm0.0\odot$ & $1.0$ &$63.0$ \\ \hline 
finetune\_final\_layer & $0.85\pm0.01\odot$ & $100863.0\pm0.0\ominus$ & $0.0\pm0.0\odot$ & $1.0$ &$63.0$ \\ \hline 
random & $0.63\pm0.33\odot$ & $3757066.73\pm3440392.9\odot$ & $0.0\pm0.0\odot$ & $5.67$ &$60.87$ \\ \hline 
few\_shot & $0.82\pm0.02\odot$ & $324094.33\pm40026.81\odot$ & $4229.97\pm512.58\oplus$ & $6.2$ &$63.0$ \\ \hline 
grad\_norm & $0.83\pm0.01\odot$ & $326880.13\pm22344.09\odot$ & $951.56\pm35.8\odot$ & $6.27$ &$63.0$ \\ \hline 
synflow & $0.83\pm0.02\odot$ & $336076.73\pm27354.67\odot$ & $1817.42\pm12.48\oplus$ & $6.47$ &$63.0$ \\ \hline 
$\phi$-score & $0.81\pm0.01\oplus$ & $338345.33\pm37314.0\odot$ & $1707.05\pm37.57\odot$ & $6.4$ &$63.0$ \\ \hline 
jacob\_cov & $0.84\pm0.01\odot$ & $347068.47\pm30656.88\odot$ & $673.58\pm19.3\odot$ & $6.2$ &$63.0$ \\ \hline 
spectral & $0.84\pm0.01\odot$ & $357978.93\pm31536.53\odot$ & $288.78\pm13.05\odot$ & $6.53$ &$63.0$ \\ \hline 
\end{tabular}}
\caption{Comparisons of different neural architecture search approaches for identifying the location and the structure of the adaptors for transfer to a new domain.}
\label{tab:zsnas}
\end{table*}

\subsection{OOD Detection and Dynamic Data Routing with Multiple Sets of Adaptors/Reconfigurators}

In this section, we look at how the model performs when there are two sets of adaptors/reconfigurators for two distinct domains, and the data must appropriately routed to the correct adaptor set. We consider five metrics: 1) accuracy of applying the adaptors from the second domain to test data from the second domain to verify that the model learns strong models on the new domain despite using a frozen encoder model, 2) applying the adaptors from the second domain to OOD test data from the first domain to see if there is some inherent robustness to noise from distribution shift in the model, 3) the macro-F1-measure, which measures how well the data is routed to the appropriate adaptor set, 4) ``combined accuracy routing score'', which is the accuracy of jointly routing to the correct set of adaptors and then making correct predictions, and the 5) ``combined accuracy shared classes score'', which is the accuracy of predicting the correct class even when routed to the wrong adaptor (if the domains share a class set). Results are shown in Table \ref{tab:zsnas2}. We observe:
\begin{itemize}
    \item Despite using a frozen encoder and a set of shallow adaptors, the model performs well on new domain data.
    \item Surprisingly, the adaptors of the new domain handle OOD samples from the original domain extremely well ($>50\%$ accuracy) when classes are shared despite not being trained on them. We expect this is because the adaptor was originally trained on the first domain, HDC is known to be robust to noise, and the learned HD classifiers can be thought of as a heavily regularized learning scheme (ensemble of many weak classifiers).
    \item The routing mechanism seems generally effective.
    \item The imperfections in routing do cause a noticeable drop in accuracy, but accuracy is still at acceptable levels considering the difficulty of the problem.
    \item Because of the strong robustness to noise due to distribution shift, even when the data samples are incorrectly routed, when classes are shared, the model can often still succeed in assigning the correct classes.
\end{itemize}

\begin{table*}[ht!]
\centering
\resizebox{\textwidth}{!}{
\begin{tabular}{|c|c|c|c|c|c|}
\hline
Setting & Accuracy\_New-to-New & Accuracy\_New-to-Old & Routing\_macro-F1 & Combined\_Accuracy\_Routing & Combined\_Accuracy\_Shared\_Classes \\ \hline 
office\_31: amazon : webcam & $0.97\pm0.01$ &$0.65\pm0.02$ &$0.79\pm0.02$ &$0.71\pm0.02$ &$0.83\pm0.02$ \\ \hline 
office\_31\_split\_classes: amazon\_split\_1 : amazon\_split\_2 & $0.9\pm0.02$ &N/A &$0.85\pm0.02$ &$0.78\pm0.02$ &N/A \\ \hline 
office\_home\_disjoint: real\_disjoint : clipart\_disjoint & $0.73\pm0.06$ &N/A &$0.72\pm0.04$ &$0.59\pm0.03$ &N/A \\ \hline 
pacs: photo : sketch & $0.84\pm0.01$ &$0.53\pm0.02$ &$0.75\pm0.17$ &$0.7\pm0.15$ &$0.8\pm0.12$ \\ \hline 
\end{tabular}
}
\caption{Understanding the robustness of the learned adaptors and ability to dynamically routing the data through the appropriate adaptor sets in the 2-domain problem}
\label{tab:zsnas2}
\end{table*}

\subsection{Class-Incremental Continual Learning}

In the final set of experiments, we create a curriculum involving six tasks, each of which consists of classifying data from the DomainNet dataset. The model is fed one sample at a time, determines which set of adaptors to send the data to, and makes predictions regarding the class label and likelihood of the sample being OOD. The selected adaptor and OOD scores are stored over a window of 50 time steps. When at least 60\% of the last 50 samples have an OOD score of 0.7 or greater, the model triggers a learning phase wherein 1000 samples are collected and labeled by an oracle. The model runs the ZS-NAS to identify the locations and structures of the adaptors, and the adaptor set is trained on the collected data. Every 2000 samples, the task changes (model has no prior knowledge of when tasks change). Each task appears twice, and the ordering of the curriculum is random. We measure the moving average over the last 50 samples as the performance metric. We compare three problem formulations: 1) an upper bound on performance where the model is given knowledge of the current task, 2) a ``slow'' dynamic routing algorithm where the model determines which set of adaptors to send the current sample to based on a majority vote using ID scores from the last 50 samples, and 3) an instantaneous dynamic routing algorithm where each data sample is treated independently, and the best adaptor is selected greedily per sample.

\begin{figure*}[t]
{\centering
\includegraphics[width=\linewidth]{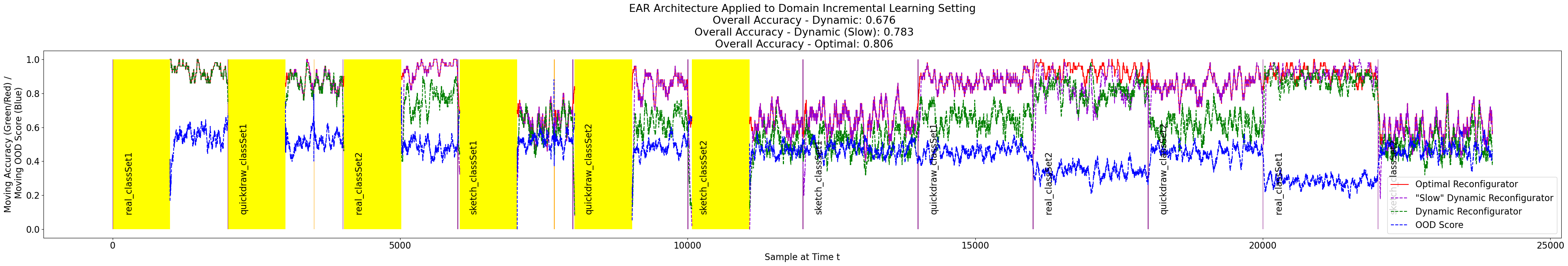}
\caption{Showing continual learning on a curriculum of six tasks repeated twice. Purple lines denote task changes. Yellow bars denote data collection and training periods. Orange lines denote misfires of detecting shifts in the distribution. All results are moving averages over the last 50 samples, reset after each training period.}
\label{fig:lifelong_learning}
}
\end{figure*}

The evaluation curves and overall accuracy of the different routing mechanisms appear in Fig. \ref{fig:lifelong_learning} where we observe:
\begin{itemize}
\item The model exhibits the successful continual learning.
\item The slow routing algorithm performs very similarly to the optimal routing algorithm, but always has a time-delay penalty after the task change due to needing to gather enough evidence to switch its current adaptor set.
\item The instantaneous routing algorithm is able to immediately correct for the task change, but it makes more mistakes about the task identity, resulting in lower overall performance compared to the ``slow'' routing algorithm.
\item The model correctly learns adaptors for the six tasks (with some early misfires). Updates are generally triggered soon after task changes.  
\item Once the six adaptors are learned, OOD scores remain low and the update phases no longer trigger.
\end{itemize}

\section{Conclusion and Future Work}
We presented a novel framework for efficient and rapid model adaptation on resource-constrained devices using the EAR architecture. We presented novel technical contributions related to progressive NNs for continual learning, deep HDC-based OOD detection, and spectral analysis-based ZS-NAS. We found that the learned HD-based approaches outperformed the standard cross-entropy-trained and deep nearest neighbor-based approaches with respect to ID accuracy by \textasciitilde 2-7\% across all datasets, demonstrating the discriminative power of the deep HD classifier. Similarly, on the OOD detection task, the learned HD model generally matched or exceeded the performance of all baseline methods. Our spectral method for ZS-NAS was shown to be 2-7x faster for heuristic evaluation
than other ZS-NAS methods while discovering in low-parameter architectures that achieved equal or better performance on the downstream classification and OOD tasks. Interestingly, in the domain-incremental setting (different modalities, same classes), the EAR architecture is shown to be extremely robust to shifts in modality, achieving high classification accuracy even when routed to the incorrect set of adaptors. Finally, in the class-incremental continual learning setting, the proposed approach effectively identifies task changes and the ``slow'' dynamic routing mechanism achieves performance close to the optimal routing mechanism (78.3\% overall accuracy vs 80.6\%).

However, there are still some components of our approach that prevent the system from being fully autonomous. Principally, the current approach requires an oracle i) to verify that the domain has truly changed and ii) to subsequently label data for the new domain. The addition of a pseudo-labeling mechanism would reduce the need for an oracle. Our proposed approach also assumes the device running the model has no bound on memory; i.e., our model continually grows new sets of adaptors. We need a mechanism for not only growing adaptors, but updating and pruning existing adaptors when the data distribution shifts to handle devices with limited memory.

\bibliographystyle{IEEEtran}
\bibliography{bibliography}

\begin{thebibliography}{10}
\providecommand{\url}[1]{#1}
\csname url@samestyle\endcsname
\providecommand{\newblock}{\relax}
\providecommand{\bibinfo}[2]{#2}
\providecommand{\BIBentrySTDinterwordspacing}{\spaceskip=0pt\relax}
\providecommand{\BIBentryALTinterwordstretchfactor}{4}
\providecommand{\BIBentryALTinterwordspacing}{\spaceskip=\fontdimen2\font plus
\BIBentryALTinterwordstretchfactor\fontdimen3\font minus
  \fontdimen4\font\relax}
\providecommand{\BIBforeignlanguage}[2]{{%
\expandafter\ifx\csname l@#1\endcsname\relax
\typeout{** WARNING: IEEEtran.bst: No hyphenation pattern has been}%
\typeout{** loaded for the language `#1'. Using the pattern for}%
\typeout{** the default language instead.}%
\else
\language=\csname l@#1\endcsname
\fi
#2}}
\providecommand{\BIBdecl}{\relax}
\BIBdecl

\bibitem{de2021continual}
M.~De~Lange, R.~Aljundi, M.~Masana, S.~Parisot, X.~Jia, A.~Leonardis,
  G.~Slabaugh, and T.~Tuytelaars, ``A continual learning survey: Defying
  forgetting in classification tasks,'' \emph{IEEE TPAMI}, 2021.

\bibitem{van2022three}
G.~M. van~de Ven, T.~Tuytelaars, and A.~S. Tolias, ``Three types of incremental
  learning,'' \emph{Nature Machine Intelligence}, 2022.

\bibitem{kanerva2009hyperdimensional}
P.~Kanerva, ``Hyperdimensional computing: An introduction to computing in
  distributed representation with high-dimensional random vectors,''
  \emph{Cognitive computation}, 2009.

\bibitem{mellor2021neural}
J.~Mellor, J.~Turner, A.~Storkey, and E.~J. Crowley, ``Neural architecture
  search without training,'' in \emph{ICML}.\hskip 1em plus 0.5em minus
  0.4em\relax PMLR, 2021.

\bibitem{rusu2016progressive}
A.~A. Rusu, N.~C. Rabinowitz, G.~Desjardins, H.~Soyer, J.~Kirkpatrick,
  K.~Kavukcuoglu, R.~Pascanu, and R.~Hadsell, ``Progressive neural networks,''
  \emph{arXiv:1606.04671}, 2016.

\bibitem{fayek2020progressive}
H.~M. Fayek, L.~Cavedon, and H.~R. Wu, ``Progressive learning: A deep learning
  framework for continual learning,'' \emph{Neural Networks}, vol. 128, 2020.

\bibitem{chen2018lifelong}
Z.~Chen and B.~Liu, ``Lifelong machine learning,'' \emph{Synthesis Lectures on
  AI and ML}, 2018.

\bibitem{yoon2017lifelong}
J.~Yoon, E.~Yang, J.~Lee, and S.~J. Hwang, ``Lifelong learning with dynamically
  expandable networks,'' \emph{arXiv:1708.01547}, 2017.

\bibitem{hung2019compacting}
C.-Y. Hung, C.-H. Tu, C.-E. Wu, C.-H. Chen, Y.-M. Chan, and C.-S. Chen,
  ``Compacting, picking and growing for unforgetting continual learning,''
  \emph{NeurIPS}, 2019.

\bibitem{li2019learn}
X.~Li, Y.~Zhou, T.~Wu, R.~Socher, and C.~Xiong, ``Learn to grow: A continual
  structure learning framework for overcoming catastrophic forgetting,'' in
  \emph{ICML}.\hskip 1em plus 0.5em minus 0.4em\relax PMLR, 2019.

\bibitem{yang2022robust}
G.~Yang, C.~S.~Y. Wong, and R.~Savitha, ``Robust continual learning through a
  comprehensively progressive bayesian neural network,''
  \emph{arXiv:2202.13369}, 2022.

\bibitem{yang2021generalized}
J.~Yang, K.~Zhou, Y.~Li, and Z.~Liu, ``Generalized out-of-distribution
  detection: A survey,'' \emph{arXiv:2110.11334}, 2021.

\bibitem{markou2003novelty}
M.~Markou and S.~Singh, ``Novelty detection: a review—part 1: statistical
  approaches,'' \emph{Signal processing}, 2003.

\bibitem{manevitz2001one}
L.~M. Manevitz and M.~Yousef, ``One-class svms for document classification,''
  \emph{JMLR}, 2001.

\bibitem{pena2001multivariate}
D.~Pe{\~n}a and F.~J. Prieto, ``Multivariate outlier detection and robust
  covariance matrix estimation,'' \emph{Technometrics}, 2001.

\bibitem{breunig2000lof}
M.~M. Breunig, H.-P. Kriegel, R.~T. Ng, and J.~Sander, ``Lof: identifying
  density-based local outliers,'' in \emph{ACM SIGMOD International Conference
  on Management of Data}, 2000.

\bibitem{liu2008isolation}
F.~T. Liu, K.~M. Ting, and Z.-H. Zhou, ``Isolation forest,'' in
  \emph{ICDM}.\hskip 1em plus 0.5em minus 0.4em\relax IEEE, 2008.

\bibitem{liu2012isolation}
------, ``Isolation-based anomaly detection,'' \emph{ACM TKDD}, 2012.

\bibitem{hendrycks2016baseline}
D.~Hendrycks and K.~Gimpel, ``A baseline for detecting misclassified and
  out-of-distribution examples in neural networks,'' \emph{arXiv:1610.02136},
  2016.

\bibitem{sastry2020detecting}
C.~S. Sastry and S.~Oore, ``Detecting out-of-distribution examples with gram
  matrices,'' in \emph{ICML}.\hskip 1em plus 0.5em minus 0.4em\relax PMLR,
  2020.

\bibitem{wilson2023hyperdimensional}
S.~Wilson, T.~Fischer, N.~S{\"u}nderhauf, and F.~Dayoub, ``Hyperdimensional
  feature fusion for out-of-distribution detection,'' in \emph{WACV}, 2023.

\bibitem{sun2022out}
Y.~Sun, Y.~Ming, X.~Zhu, and Y.~Li, ``Out-of-distribution detection with deep
  nearest neighbors,'' in \emph{ICML}.\hskip 1em plus 0.5em minus 0.4em\relax
  PMLR, 2022.

\bibitem{thomas2021theoretical}
A.~Thomas, S.~Dasgupta, and T.~Rosing, ``A theoretical perspective on
  hyperdimensional computing,'' \emph{JAIR}, 2021.

\bibitem{kleyko2023survey}
D.~Kleyko, D.~A. Rachkovskij, E.~Osipov, and A.~Rahimi, ``A survey on
  hyperdimensional computing aka vector symbolic architectures, part i: Models
  and data transformations,'' \emph{ACM Computing Surveys}.

\bibitem{zhuang2020comprehensive}
F.~Zhuang, Z.~Qi, K.~Duan, D.~Xi, Y.~Zhu, H.~Zhu, H.~Xiong, and Q.~He, ``A
  comprehensive survey on transfer learning,'' \emph{Proceedings of the IEEE},
  2020.

\bibitem{elsken2019neural}
T.~Elsken, J.~H. Metzen, and F.~Hutter, ``Neural architecture search: A
  survey,'' \emph{JMLR}, 2019.

\bibitem{abdelfattah2021zero}
M.~S. Abdelfattah, A.~Mehrotra, {\L}.~Dudziak, and N.~D. Lane, ``Zero-cost
  proxies for lightweight nas,'' \emph{arXiv:2101.08134}, 2021.

\bibitem{chen2021bench}
H.~Chen, M.~Lin, X.~Sun, and H.~Li, ``Nas-bench-zero: A large scale dataset for
  understanding zero-shot neural architecture search,'' 2021.

\bibitem{lee2018snip}
N.~Lee, T.~Ajanthan, and P.~H. Torr, ``Snip: Single-shot network pruning based
  on connection sensitivity,'' \emph{arXiv:1810.02340}, 2018.

\bibitem{wang2020picking}
C.~Wang, G.~Zhang, and R.~Grosse, ``Picking winning tickets before training by
  preserving gradient flow,'' \emph{arXiv:2002.07376}, 2020.

\bibitem{turner2019blockswap}
J.~Turner, E.~J. Crowley, M.~O'Boyle, A.~Storkey, and G.~Gray, ``Blockswap:
  Fisher-guided block substitution for network compression on a budget,''
  \emph{arXiv:1906.04113}, 2019.

\bibitem{tanaka2020pruning}
H.~Tanaka, D.~Kunin, D.~L. Yamins, and S.~Ganguli, ``Pruning neural networks
  without any data by iteratively conserving synaptic flow,'' \emph{NeurIPS},
  2020.

\bibitem{lin2021zen}
M.~Lin, P.~Wang, Z.~Sun, H.~Chen, X.~Sun, Q.~Qian, H.~Li, and R.~Jin,
  ``Zen-nas: A zero-shot nas for high-performance image recognition,'' in
  \emph{ICCV}, 2021.

\bibitem{lin2017focal}
T.-Y. Lin, P.~Goyal, R.~Girshick, K.~He, and P.~Doll{\'a}r, ``Focal loss for
  dense object detection,'' in \emph{ICCV}, 2017.

\bibitem{scheirer2014probability}
W.~J. Scheirer, L.~P. Jain, and T.~E. Boult, ``Probability models for open set
  recognition,'' \emph{IEEE TPAMI}, 2014.

\bibitem{rinne2008weibull}
H.~Rinne, \emph{The Weibull distribution: a handbook}.\hskip 1em plus 0.5em
  minus 0.4em\relax CRC Press, 2008.

\bibitem{snoek2012practical}
J.~Snoek, H.~Larochelle, and R.~P. Adams, ``Practical bayesian optimization of
  machine learning algorithms,'' \emph{NeurIPS}, 2012.

\bibitem{bayesian}
\BIBentryALTinterwordspacing
F.~Nogueira, ``{Bayesian Optimization}: Open source constrained global
  optimization tool for {Python},'' 2014--. [Online]. Available:
  \url{https://github.com/fmfn/BayesianOptimization}
\BIBentrySTDinterwordspacing

\bibitem{stander2002robustness}
N.~Stander and K.~Craig, ``On the robustness of a simple domain reduction
  scheme for simulation-based optimization,'' \emph{Engineering Computations},
  2002.

\bibitem{chen2021understanding}
W.~Chen, X.~Gong, J.~Wu, Y.~Wei, H.~Shi, Z.~Yan, Y.~Yang, and Z.~Wang,
  ``Understanding and accelerating neural architecture search with
  training-free and theory-grounded metrics,'' \emph{arXiv:2108.11939}, 2021.

\bibitem{marsden2013eigenvalues}
A.~Marsden, ``Eigenvalues of the laplacian and their relationship to the
  connectedness of a graph,'' \emph{University of Chicago, REU}, 2013.

\bibitem{von2007tutorial}
U.~Von~Luxburg, ``A tutorial on spectral clustering,'' \emph{Statistics and
  computing}, 2007.

\bibitem{damle2019simple}
A.~Damle, V.~Minden, and L.~Ying, ``Simple, direct and efficient multi-way
  spectral clustering,'' \emph{Information and Inference: A Journal of the
  IMA}, 2019.

\bibitem{vinh2009information}
N.~X. Vinh, J.~Epps, and J.~Bailey, ``Information theoretic measures for
  clusterings comparison: is a correction for chance necessary?'' in
  \emph{ICML}, 2009.

\bibitem{tan2021efficientnetv2}
M.~Tan and Q.~Le, ``Efficientnetv2: Smaller models and faster training,'' in
  \emph{ICML}.\hskip 1em plus 0.5em minus 0.4em\relax PMLR, 2021.

\bibitem{kingma2014adam}
D.~P. Kingma and J.~Ba, ``Adam: A method for stochastic optimization,''
  \emph{arXiv:1412.6980}, 2014.

\bibitem{khosla2020supervised}
P.~Khosla, P.~Teterwak, C.~Wang, A.~Sarna, Y.~Tian, P.~Isola, A.~Maschinot,
  C.~Liu, and D.~Krishnan, ``Supervised contrastive learning,'' \emph{NeurIPS},
  2020.

\bibitem{li2017deeper}
D.~Li, Y.~Yang, Y.-Z. Song, and T.~M. Hospedales, ``Deeper, broader and artier
  domain generalization,'' in \emph{ICCV}, 2017.

\bibitem{saenko2010adapting}
K.~Saenko, B.~Kulis, M.~Fritz, and T.~Darrell, ``Adapting visual category
  models to new domains,'' in \emph{ECCV}.\hskip 1em plus 0.5em minus
  0.4em\relax Springer, 2010.

\bibitem{venkateswara2017deep}
H.~Venkateswara, J.~Eusebio, S.~Chakraborty, and S.~Panchanathan, ``Deep
  hashing network for unsupervised domain adaptation,'' in \emph{CVPR}, 2017.

\bibitem{peng2019moment}
X.~Peng, Q.~Bai, X.~Xia, Z.~Huang, K.~Saenko, and B.~Wang, ``Moment matching
  for multi-source domain adaptation,'' in \emph{ICCV}, 2019.

\end{thebibliography}

\begin{IEEEbiography}{Dr. Zachary A. Daniels}{\space} is a research computer scientist with the Vision Systems Lab of SRI International's Center for Vision Technologies in Princeton, NJ. He received his bachelor's in computer science with a minor in cognitive science from Lehigh University in 2014. He received his doctorate in computer science with a focus on machine learning, computer vision, and artificial intelligence in 2020 from Rutgers University.
\end{IEEEbiography}

\begin{IEEEbiography}{Dr. David Zhang}{\space} is a Senior Technical Manager in the Vision Systems Lab of the Center for Vision Technologies at SRI International with experience in algorithm and embedded software development. He has expertise in computer vision and machine learning algorithms, with a focus on edge computing and video surveillance, tracking, and enhancements in degraded visual environments. He received his PhD in Physics from The Pennsylvania State University in 2001. 
\end{IEEEbiography}

\end{document}